\documentclass[journal]{IEEEtran}
\IEEEoverridecommandlockouts

\usepackage{amsmath,amsfonts}
\usepackage{array}
\usepackage{textcomp}
\usepackage{url}
\usepackage{verbatim}
\usepackage{graphicx}
\usepackage{cite}
\usepackage{float}
\usepackage{bm}
\usepackage{color}
\usepackage{multirow}
\usepackage{multicol}
\usepackage{subcaption}
\usepackage{hyperref}
\usepackage{cite}
\usepackage{graphicx}
\usepackage{subcaption}
\usepackage{tabularx}
\usepackage{dblfloatfix}
\usepackage{comment}
\usepackage{algorithm}
\usepackage{algorithmicx}
\usepackage[noend]{algpseudocode}
\usepackage{mathtools}
\usepackage{amsmath,amsfonts,amssymb}
\IEEEoverridecommandlockouts 
\usepackage[dvipsnames]{xcolor}

\graphicspath{ {./figures/} }

\title{\LARGE \bf
C3D: Cascade Control with Change Point Detection and Deep Koopman Learning for Autonomous Surface Vehicles}

\author{Jianwen Li$^{1}$, Hyunsang Park$^{1}$, Wenjian Hao$^{1}$, Lei Xin$^{1}$, Jalil Chavez-Galaviz$^{1}$,  Ajinkya Chaudhary$^{1}$, Meredith Bloss$^{1}$, Kyle Pattison$^{2}$, Christopher Vo$^{2}$, Devesh Upadhyay$^{2}$, Shreyas Sundaram$^{1}$, Shaoshuai Mou$^{1}$, Inseok Hwang$^{1}$, and Nina Mahmoudian$^{1}$

\thanks{$^{1}$ College of Engineering, Purdue University, West Lafayette, IN, USA. }
\thanks{$^{2}$ Saab, Inc., USA. }
\thanks{* This work has been submitted to the IEEE for possible publication. Copyright may be transferred without notice, after which this version may no longer be accessible.}
}

\begin{document}

\maketitle
\thispagestyle{empty}
\pagestyle{empty}


\begin{abstract}
In this paper, we discuss the development and deployment of a robust autonomous system capable of performing various tasks in the maritime domain under unknown dynamic conditions. We investigate a data-driven approach based on modular design for ease of transfer of autonomy across different maritime surface vessel platforms. The data-driven approach alleviates issues related to \textit{a} priori identification of system models that may become deficient under evolving system behaviors and/or shifting, unanticipated, environmental influences.  Our proposed learning-based platform comprises a deep Koopman system model and a change point detector that provides guidance on domain shifts prompting relearning under severe exogenous and endogenous perturbations. Motion control of the autonomous system is achieved via an optimal controller design. The Koopman linearized model naturally lends itself to a linear–quadratic regulator (LQR) control design. We propose the C3D control architecture “Cascade Control with Change Point Detection and Deep Koopman Learning”. The framework is verified in station keeping task on an ASV in both simulation and real experiments. The approach achieved at least $\textbf{13.9\%}$ improvement in mean distance error in all test cases compared to the methods that do not consider system changes.
\end{abstract}


\maketitle

\renewcommand\thefootnote{}
\renewcommand\thefootnote{\fnsymbol{footnote}}
\setcounter{footnote}{1}

\section{INTRODUCTION}

Autonomous Surface Vessels (ASVs) are maritime domain vessels equipped with various sensors, navigation systems, and control mechanisms, allowing them to perform tasks autonomously. The applications of autonomous surface vessels are diverse and continually expanding and include applications such as maritime surveillance, oceanographic research and mapping, search and rescue, anti-submarine warfare, offshore energy support, communications relays, autonomous replenishment, cargo transport, etc. In addition, ASVs can also enhance efficiency, reduce costs, and mitigate risks associated with human involvement in various marine operations. With the rapid advances in edge computing and embedded real-time capabilities, the range of applications for ASVs is expected to expand further. However, deploying efficient ASVs that are robust, self-healing, safe, and reliable introduces several engineering challenges. Additionally, rapid transfer of autonomy from sim-to-prototype-to-real (Sim2P2R) requires generalizability capabilities that inherently force systems to be capable of on-device learning/relearning \cite{hofer2021sim2real}. These requirements are made increasingly difficult under unanticipated damage and modifications to the system, which may occur in complex and dynamic environments without warning and may not have any prior analytic representation. In this sense,  damage and modifications can be tackled either via system dynamic updates and/or through disturbance rejection control methods including adaptation. Adaptive controllers rely on a parametric design and mechanisms for tuning these parameters. The effectiveness of adaptation methods is, however, often limited to bounded drifts in low-dimensional systems \cite{ioannou1996robust}.   System dynamic update techniques like state observers, intrusive system identification, etc. are often used.  However, successful convergence to the drifted system state is often highly dependent on the representation afforded by the reduced-order models of the system dynamics that typically ignore higher-order effects. Model-free methods such as active disturbance rejection control (ADRC) have attempted to alleviate some of these issues \cite{han2009pid}. In the autonomous setting, especially ASVs, properties like self-healing become very relevant given that these systems are expected to operate under severe conditions including under adversarial disablement.

\begin{figure}[!t]
\centering
    \includegraphics[width=0.48\textwidth]{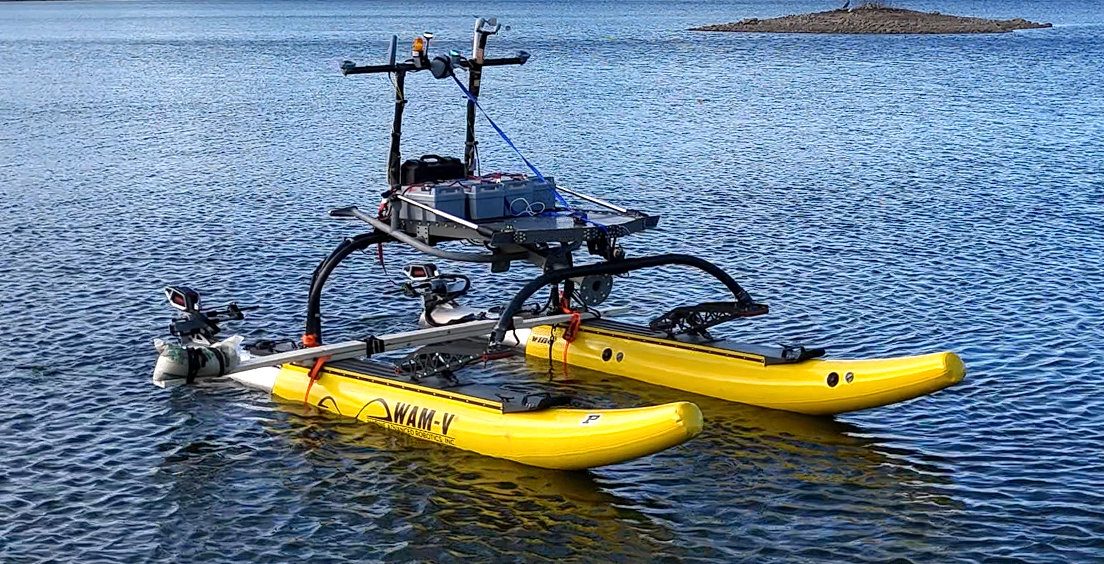}
\caption{WAM-V 16 ASV with an unbalance load performing station keeping in Lake Harner, IN}
\label{fig:wamv_real_nom}
\end{figure}
In such scenarios, learning-based methods may offer an advantage over classical methods. As such, we experiment with a deep Koopman based control construct with active online relearning of the altered system dynamic under a spectrum of unanticipated disturbances. The full scope of this work includes an ASV autonomy stack covering diverse ASV applications such as station keeping and payload transfer and a set of maneuvers including optimal path planning under nominal as well as deviated conditions that include system modifications and drifts. Station keeping allows the ASV to maintain a specific position and heading for some predefined duration. An ASV performing station keeping is shown in Fig. \ref{fig:wamv_real_nom}. The ability to perform such maneuvers significantly enhances the autonomy of ASVs, expanding their utility across diverse missions such as biological surveys, maritime rescues \cite{qu2022nonlinear, sarda2016station}, Launch and Recovery (LAR) operations \cite{sarda2018launch}, and docking in moderate water conditions.



Extensive research efforts have focused on the study of the station keeping, for instance, developing an ASV for autonomous missions in challenging surf zone areas, utilizing stereo vision to counter wave disturbance \cite{chen2013station}. Further studies have explored differential drive ASVs with azimuth propulsion and different control techniques such as PD, backstepping, and sliding mode controllers, including wind disturbance compensation \cite{sarda2016station, sarda2018launch}. Other work has centered its attention on addressing the station keeping problem for underactuated vehicles under moderate wind conditions with relative success \cite{pereira2008experimental}. In more recent studies, simulations have been used to compare the implementation of a nonlinear controller for station keeping and highlight its advantages over a reinforcement learning controller \cite{sinisterra2020nonlinear}, while others have utilized simulations to assess the performance of different controllers for station keeping on underactuated ASVs, focusing on the energy efficiency \cite{oh2020comparison}. Other work has been proposed using a cascaded controller for surge, sway, and yaw control, reducing actuation energy by computing the average disturbance direction \cite{qu2022nonlinear}. Finally, a data-driven Model Predictive Control (MPC) method is proposed to model the nonlinear ASV systems under wind disturbances and perform station keeping with an underactuated vehicle \cite{chavez2023asv}. All of this work fails to consider the influence of change in system dynamics online. 

While there is a long history on online change point detection (CPD) \cite{tartakovsky2012efficient, flynn2019change, chang2018kernel}, only a limited number of works are applicable to dynamical systems. Compared to standard models considered in the classical CPD settings, the key difference in a dynamical system is that the data points (e.g., measured states) are not independent. Furthermore, to obtain theoretical guarantees such as false alarm rate, a key assumption made in existing works is that perfect knowledge of model parameters before or after the change is available \cite{ lai1998information, banerjee2015data}. The recent work \cite{xin2023online} proposed an online CPD test for unknown dynamical systems and an associated data-dependent threshold that enables the user to achieve a pre-specified upper bound on the false alarm probability. However, the theoretical guarantees only hold for linear systems. In this work, we propose a composite online CPD algorithm that leverages the ideas in \cite{xin2023online} and an additional tracking error-based statistic, which enables one to detect changes for nonlinear systems without known model parameters in advance.

 
In this paper, we present results from the current stage of this work. We present results from optimal waypoint following and station keeping for both nominal as well as for a system under various degrees of disablement.  We also consider the non-holonomic problem.  Our autonomy solution, \textbf{C}ascade \textbf{C}ontrol with \textbf{C}hange Point Detection and \textbf{D}eep Koopman Learning (C3D), has three modules:  
\begin{itemize}
\item The system dynamics are defined using a deep Koopman model \cite{hao2024deep}. The Koopman approach allows a globally linear representation of a non-linear dynamical system in a high dimensional lifted space. A deep network is used to learn a finite dimensional representation of the Koopman operator.  This approach provides two benefits: first, it allows the use of any linear optimal control, and second, quick online re-learning makes it possible to update the system state space under system drift. 
\item A change point detector (CPD) that can detect and quantify system drift. The CPD acts as a trigger for the relearning sequence of the deep Koopman representation (DKR).     
\item For control, we use a cascaded LQR acting on the Koopman approximation of the current system dynamics.\\
\end{itemize}
 
For the initial development, the Wave Adaptive Modular Vehicle (WAM-V) was intentionally underactuated with only two degrees of freedom (fixed port and starboard thrusters) by disabling the rudder mechanism for steering. This made the WAM-V non-holonomic, however, we were able to demonstrate reasonable steering via differential drive. We first developed the framework in a Gazebo simulation environment and performed several comparative investigations including benchmarking against a simple PID controller.  Following this, the C3D was deployed on the WAM-V in a sim-to-real transfer and tested under real-world test conditions including actively induced system perturbations.

The rest of the paper is organized as follows: In section II, the station keeping problem statement is defined along with a 3-DOF reference model of an ASV, and the proposed C3D architecture is presented. In Section III we provide the details of the simulation validation and describe validation on the commercial off-the-shelf WAM-V 16 ASV. Finally, conclusions and future work are discussed in Section IV.\\

\section{METHODOLOGY}\label{sec:methodology}
 
\subsection{WAM-V 16 and the Simulator }
As described in \cite{fossen2011handbook}, a mathematical model of an ASV can be described as follows:
 
\begin{align}
    \dot{\eta} &= R(\psi){\nu} \label{eq:asv_kin}\\
     M\dot{ \nu} + {C}({\nu}){\nu} + {D}({\nu}){\nu} &= {\tau} +  \tau_w \label{eq:asv_dyn_model}
\end{align}

where $ \eta = [x,y,\psi]^{T}$ defines the ASV's pose in an inertial coordinate system. The speed vector in the body fixed frame $\nu=[u,v,r]^{T}$ consists of the linear velocities $(u,v)$ in the surge and sway directions, and  $r$ is the rotation velocity about the Z-axis. The thrust vector $ \tau$ contains the force and moment produced by the port and starboard trolling motor commands $ a=[a_p,a_s]$, respectively. The disturbance $\tau_w$ is produced by the wind and wave. The matrix $R(\psi)$ is the rotation matrix and is used as a transform between the body fixed frame to the earth (inertial) fixed frame. $M$, $ C(\nu)$, and $D(\nu)$ represent the inertia matrix, the Coriolis and centripetal force matrix, and the damping matrix, respectively.


\subsection{System Architecture}\label{system_arch}

\begin{figure}[!t]
\centering
\includegraphics[width=0.5\textwidth]{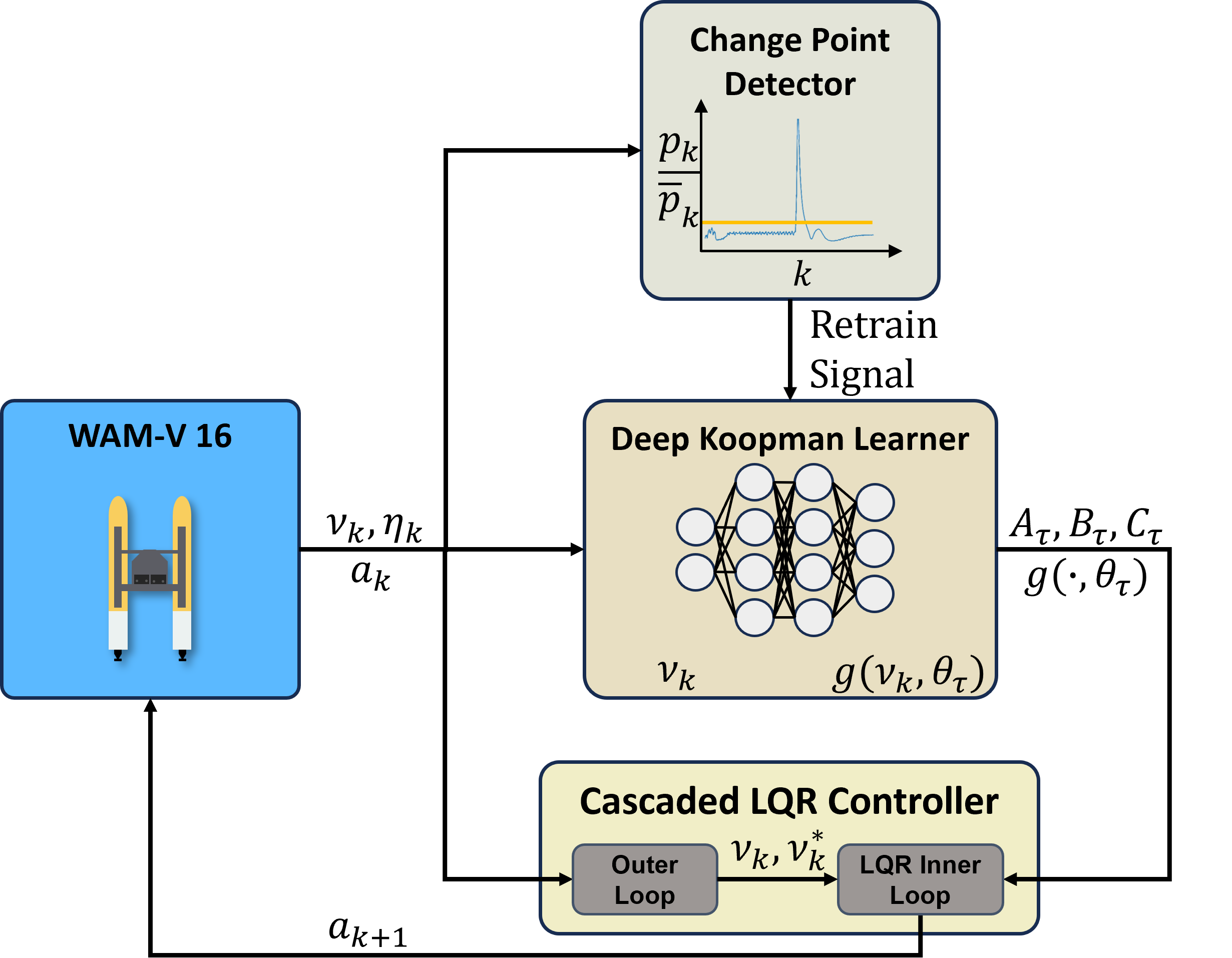}
\caption{System architecture. At every time step, the vehicle sends the current state $s_t$ and control command $a_t$ to the change point detector, deep Kooman learner, and cascade controller. The inner loop of the cascade controller uses the newest matrices $A_\tau, B_\tau, C_\tau$ and Koopman operator $g(\cdot,\theta_\tau)$ to track the target velocity $\nu_k^*$ generated by the outer loop. 
}
\label{fig:sys_arch}
\end{figure}

Fig.\ref{fig:sys_arch} shows the architecture of the C3D system. Algorithm \ref{algorithm_1} summarizes a systematic procedure to use C3D to solve control problems for time-varying systems. C3D consists of three submodules: the change point detector, the deep Koopman learner and the cascade controller.  The submodules are explained in the following subsections.

\begin{algorithm}[h]
    \caption{Cascade Control with change point Detection and
Deep Koopman Learning}\label{algorithm_1}
\begin{algorithmic}
\State Collect dataset $\mathcal{D}_n$ using random walk
\State Train nominal deep Koopman operator $g(\cdot,\theta_\tau)$ 
\State Generate matrices $A_\tau,B_\tau,C_\tau$ for the nominal model
\State Initialize cascade controller and change point detector
\Repeat
\State Observe state $\nu_k$ and $\eta_k$
\State Calculate target velocity $\nu_k^*$ using the outer loop 
\State Lift velocity $\nu_k$ and $\nu_k^*$ to $h_k$ and $h_k^*$
\State Calculate control command $u_k$ using inner loop 
\State Detect change points  
\If{Change point is detected}
    \State Collect dataset $\mathcal{D}_{retrain}$ using random walk
    \Repeat
        \State Retrain deep Koopman operator $g(\cdot,\theta)$
        \State Generate matrices $A_\tau, B_\tau, C_\tau$ for the new model
        \State Validate the cascade controller in simulation 
    \Until Convergence
    \State Update $g(\cdot,\theta_\tau)$ and $A_\tau,B_\tau,C_\tau$ for cascade controller 
    
\EndIf

\Until Maximum test steps
\end{algorithmic}
\end{algorithm}

\subsubsection{Deep Koopman Operator}

Motivated by the recent work \cite{hao2024deep} which proposes to use the Koopman operator to approximate nonlinear time-varing systems using deep neural networks, we seek to build an estimated discrete-time dynamical system based on the observed system states-inputs data pairs, of which the dynamics are with system state $g(\nu_k, \theta_\tau): \mathbb{R}^n \rightarrow \mathbb{R}^{r}$ parameterized by $\theta_{\tau}\in\mathbb{R}^{p}$, control input $a(t)\in\mathbb{R}^m$, and constant matrices $A_\tau \in \mathbb{R}^{r \times r}$, $B_{\tau} \in \mathbb{R}^{r \times m}$, $C_\tau \in \mathbb{R}^{n \times r}$ such that for $k_\tau\leq k<k_\tau+\beta_\tau$, and any given $g(\nu_{k_\tau},\theta_\tau), a_{k_\tau}$, the following holds approximately.
\begin{align} 
    g(\nu_{k+1},\theta_\tau) &= A_\tau g(\nu_k,\theta_\tau) + B_\tau a_{k}, \label{eqq1} \\ \nu_{k+1} &= C_\tau g(\nu_{k+1},\theta_\tau), \label{eqq2} 
\end{align}
Here, note that for any data batch $\mathcal{B}_\tau$, matrices $A_\tau$, $B_\tau$, $ C_\tau$ remain constant for $k\in\mathbb{K}_\tau$. For notation brevity, we denote $g(\cdot, \theta_\tau)$, $A_\tau$, $B_\tau$, $C_\tau$ achieved from \eqref{eqq1}-\eqref{eqq2} as a set $\mathcal{K}_{\mathcal{B}_\tau}$ described as follows. \begin{equation}\label{eq_dkr}
\mathcal{K}_{\mathcal{B}_\tau}\coloneqq \{g(\cdot, \theta_\tau), A_\tau, B_\tau, C_\tau\},
\end{equation} 
which is referred to as the deep Koopman representation (DKR) throughout the rest of this paper.
 
One way to achieve the DKR defined in \eqref{eq_dkr} is to obtain an optimal $\theta_\tau$ by minimizing the following objective problem based on the data batch $\mathcal{B}_{\tau}$.
\begin{equation}\label{eq_obj0}
    \begin{aligned}
        \theta_{\tau}^*= \arg\min_{\theta_\tau\in\mathbb{R}^{p_i}} \{ w \mathbf{L}_1(A_{\tau},B_\tau,\theta_\tau) + (1-w)\mathbf{L}_2(C_\tau,\theta_\tau)\}, 
    \end{aligned}
\end{equation}
where $0<w<1$ denotes the user-defined weight, $\mathbf{L}_1$ and $\mathbf{L}_2$ are loss functions to approximate \eqref{eqq1} and \eqref{eqq2} respectively and they are defined as:
\begin{equation} \label{pro1}
\begin{aligned}
\mathbf{L}_1(A_{\tau}, B_\tau,\theta_\tau)  = &\frac{1}{\beta_{\tau}}\sum_{k=k_\tau}^{k_\tau+\beta_{ \tau}-1}\parallel g(\nu_{k+1}, \theta_\tau) -(A_{\tau} g(\nu_k, \theta_\tau) \\
    + & B_\tau a_{k})\parallel^2
\end{aligned}
\end{equation}
and
\begin{equation}\label{pro11}
    \mathbf{L}_2(C_\tau,\theta_\tau)= \frac{1}{\beta_{\tau}} \sum_{k=k_\tau}^{k_\tau+\beta_{\tau}-1}\parallel \nu_k - C_\tau g(\nu_k, \theta_\tau)\parallel^2.
\end{equation} 

To proceed, one needs to partition the observed $\nu_{k}$ and $a_{k}$ as the following data batches \begin{equation}\label{eq_batch}
    \mathcal{B}_\tau=\{\nu_k,a_k:  k\in \mathbb{K}_\tau \}, \quad \tau=0,1,2,\cdots,
\end{equation} where  $\mathbb{K}_\tau=\{k_\tau,k_\tau+1,k_\tau+2,\cdots,k_{\tau}+\beta_\tau\}$ denotes a set of the ordered sampling instances for the $\tau$-th data batch $\mathcal{B}_{i,\tau}$ with $\beta_\tau$ positive integers such that $$k_\tau=\sum_{i=0}^{\tau-1} \beta_i,\quad \tau\geq 1, \quad k_0=0.$$
To solve \eqref{eq_obj0}, one needs to rewrite the data batches and $\mathbf{L}_1$ and $\mathbf{L}_2$ in compact forms.
To this end, we introduce the following notations:
\begin{equation}\label{xyudata}
    \begin{aligned}
    \mathbf{X}_\tau &=[\nu_{k_\tau}, \nu_{k_\tau+1},\cdots,\nu_{k_\tau+\beta_{\tau}-1}] \in \mathbb{R}^{n \times \beta_{\tau}},\\
    \bar{\mathbf{X}}_\tau &= [\nu_{k_\tau+1}, \nu_{k_\tau+2},\cdots,\nu_{k_\tau+\beta_{\tau}}]\in \mathbb{R}^{n \times \beta_{\tau}}\nonumber,\\
    \mathbf{U}_\tau &=[a_{k_\tau}, a_{k_\tau+1},\cdots,a_{k_\tau+\beta_{\tau}-1}]\in \mathbb{R}^{m \times \beta_{\tau}}.\nonumber
    \end{aligned}
\end{equation}
Thus $\mathbf{L}_{1}$ in (\ref{pro1}) and $\mathbf{L}_{2}$ in \eqref{pro11} can be rewritten as
\begin{equation}\label{dktvmins1}
\begin{aligned}
     \mathbf{L}_{1}= \frac{1}{\beta_{\tau}}\parallel\bar{\mathbf{G}}_{\tau}- (A_{\tau}\mathbf{G}_\tau + B_\tau\mathbf{U}_\tau)\parallel_F^2
\end{aligned}
\end{equation} and
\begin{equation}\label{dktvmins2}
\begin{aligned}
     \mathbf{L}_{2} = \frac{1}{\beta_{\tau}}\parallel \mathbf{X}_\tau - C_\tau\mathbf{G}_{\tau} \parallel_F^2, 
\end{aligned}
\end{equation} where
\begin{equation}\label{gstack1}
\begin{aligned}
    \mathbf{G}_\tau= [g(\nu_{k_\tau},\theta_\tau),\cdots, 
    g(\nu_{k_\tau+\beta_{\tau}-1}, \theta_\tau)] \in \mathbb{R}^{r \times \beta_{\tau}},\\
\bar{\mathbf{G}}_\tau= [g(\nu_{k_\tau+1}, \theta_\tau),\cdots,g(\nu_{k_\tau+\beta_{\tau}}, \theta_\tau)] \in \mathbb{R}^{r \times \beta_{\tau}}. 
\end{aligned}
\end{equation}
By minimizing $\mathbf{L}_{1}$ with respect to $A_{\tau}, B_\tau$ in \eqref{dktvmins1} and minimizing $\mathbf{L}_{2}$ regarding $C_\tau$ in \eqref{dktvmins2}, $A_\tau, B_\tau, C_\tau$ can be determined by $\theta_\tau$ as:
\begin{align}
    [A_{\tau}^{\theta}, B_\tau^{\theta }] &= \bar{\mathbf{G}}_\tau \begin{bmatrix} \mathbf{G}_\tau \\ \mathbf{U}_\tau \end{bmatrix}^\dagger \label{lmn},\\
  C_\tau^{\theta } &= \mathbf{X}_\tau\mathbf{G}_\tau^{\dagger}. \label{lmn1}
\end{align}
Here, to ensure the existence of a unique solution of \eqref{lmn}-\eqref{lmn1}, we need to assume that
the matrices $\mathbf{G}_\tau \in \mathbb{R}^{r\times \beta_\tau}$ defined in \eqref{gstack1} and $\begin{bmatrix} \mathbf{G}_\tau \\ \mathbf{U}_\tau \end{bmatrix}\in \mathbb{R}^{(r+m)\times \beta_\tau}$ are of full row rank.

\begin{figure}
     \centering
     \begin{subfigure}[b]{0.45\textwidth}
         \centering
         \includegraphics[width=\textwidth]{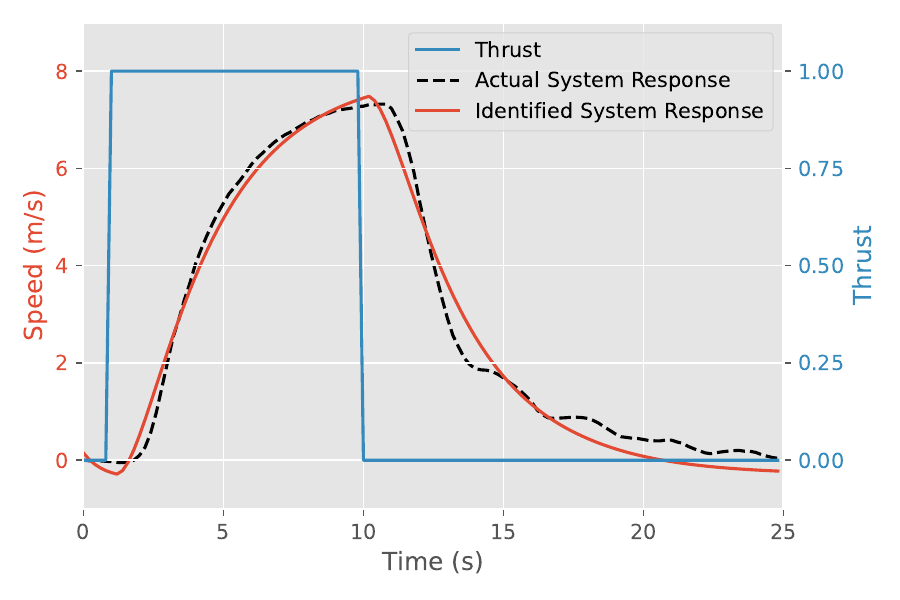}
         \caption{Input and response}
     \end{subfigure}
     \hfill
     \begin{subfigure}[b]{0.45\textwidth}
         \centering
         \includegraphics[width=\textwidth]{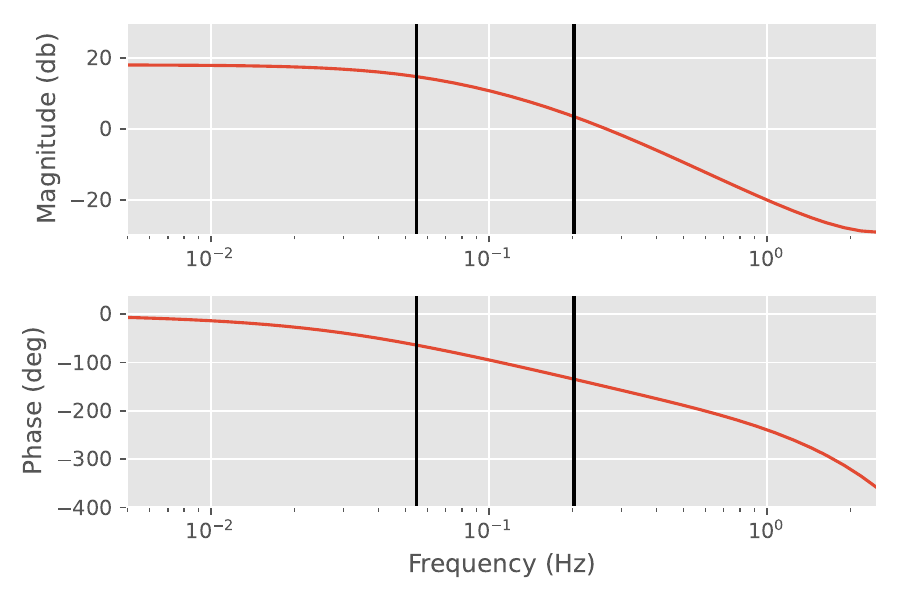}
         \caption{Bode plot}
     \end{subfigure}
        \caption{Step input response and bode plot of identified linear system}
        \label{fig:step_response}
\end{figure}

\begin{figure}
     \centering
         \includegraphics[width=0.5\textwidth]{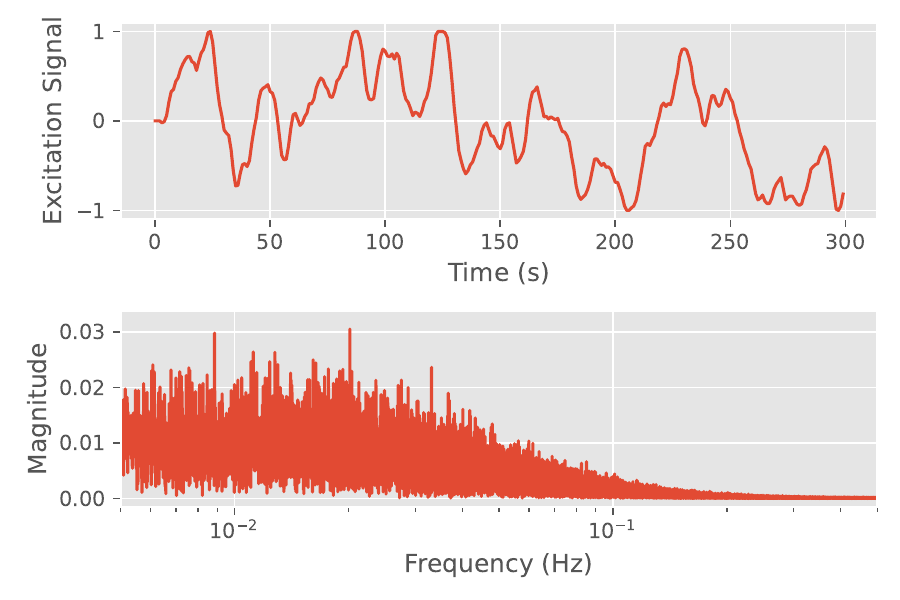}
        \caption{Example of the designed random walk signal (top) and the Fourier transform (bottom)}
        \label{fig:excitation_example}
\end{figure}

To learn the DKR online, we excite the system using a sequence of input with excitation signals, collect the state response of the system, and use the collected data to train the DKR. By carefully designing the excitation signal, we expect the size of data required to train the DKR can be significantly reduced while maintaining the accuracy of the trained DKR. Inspired by classical system identification, we design the signal such that its bandwidth can encompass that of the system. The signal should also allow us to explore a wide region of the state-input space, as the aim of the DKR is to learn a global representation of the unknown nonlinear system. In this regard, we choose random walk as the excitation signal and manually tune the parameters of the random walk such that its bandwidth includes that of the system. To obtain an estimate of the bandwidth of the system, we measured the step response (forward thrust to speed response) of the vehicle and identified a linear system with an input delay that closely fits the measured response (see Fig. \ref{fig:step_response}). Two poles were identified (indicated as the vertical lines in the Bode plot), corresponding to the system’s natural frequency and the thrusters’ natural frequency, respectively, along with an input delay of 0.4s. To mitigate the effect of thruster dynamics on the collected data, we limited the time step of the signal to 1Hz and tuned the bandwidth of the random walk to cover the bandwidth of the system, while limiting the excitation signal in higher frequencies. The excitation signal is generated from the system
\begin{equation}
    \begin{aligned}
        \zeta_{t+1} &= A_\zeta \zeta_{t} + B_\zeta w_t \\
        \xi_t &= C_\zeta \zeta_t
    \end{aligned}
\end{equation}
where $\xi_t \in \mathbb{R}$ is the excitation signal, $\zeta_t \in \mathbb{R}^{n_\zeta}$ is the internal state of the system, and $w_t \in \mathbb{R}$ is the standard gaussian noise. The parameters $A_\zeta$, $B_\zeta$ and $C_\zeta$ were tuned manually. Fig. \ref{fig:excitation_example} shows an example of the excitation signal, as well as the Fourier transformation of the signal. 

\subsubsection{Change point Detector} 

To detect changes in system dynamics, we leverage a variant of the online CPD algorithm proposed in \cite{xin2023online}. Specifically, \cite{xin2023online} requires to maintain two sliding windows (i.e., a reference window and a test window) of data at each time step. After that, the approach estimates a model using data from each window and flags a change point if the norm of the difference of the estimated models is larger than some threshold value. Below we provide an overview of the main ideas of the algorithm in \cite{xin2023online}.

Let $N_{1}\geq 2$ be the window size parameter that captures how many samples one would like to use at each time step. Let $\{(\nu_{i},u_{j}):0 \leq i \leq 2N_{1}-1,0 \leq j \leq 2N_{1}-2\}$ be the initial dataset. At each time step $k\geq 2N-1$, one collects the sample pair $(\nu_{k+1},a_{k})$ generated by the system. Define the following matrices for every time step $k\geq 2N_{1}-1$ 
\begin{equation} 
\begin{aligned}
X^{ref}_{k}=\begin{bmatrix}
\nu_{k-2N_{1}+3}&\cdots&\nu_{k-N_{1}+1}
\end{bmatrix}  \in \mathbb{R}^{n\times (N_{1}-1)},\\
\end{aligned}
\end{equation}
and 
\begin{equation} 
\begin{aligned}
X^{test}_{k}=\begin{bmatrix}
\nu_{k-N_{1}+3}&\cdots&\nu_{k+1}
\end{bmatrix}  \in \mathbb{R}^{n\times (N_{1}-1)},
\end{aligned}
\end{equation}
respectively. Denote $\xi_{k}=\begin{bmatrix} \nu_{k}^T& a_{k}^T\end{bmatrix}^{T}\in \mathbb{R}^{n+m}$ for $k\geq 0$, and further define the following matrices at time step $k$
\begin{equation} 
\begin{aligned}
&Z^{ref}_{k}=\begin{bmatrix}
\xi_{k-2N_{1}+2}&\cdots&\xi_{k-N_{1}}
\end{bmatrix} \in \mathbb{R}^{(n+m)\times (N_{1}-1)},\\
&Z^{test}_{k}=\begin{bmatrix}
\xi_{k-N_{1}+2}&\cdots&\xi_{k}
\end{bmatrix} \in \mathbb{R}^{(n+m)\times (N_{1}-1)}.
\end{aligned}
\end{equation}

Let $\hat{\Theta}_{k}^{ref}$ and $\hat{\Theta}_{k}^{test}$ be the estimated system models, which are computed by solving the following least squares problems
\begin{equation*}
\begin{aligned}
  \mathop{\min}_{\tilde{\Theta}^{ref}_{k}\in \mathbb{R}^{n\times (n+m)}} \{\|X^{ref}_{k}-\tilde{\Theta}^{ref}_{k}Z^{ref}_{k}\|^{2}_{F}+\lambda \|\tilde{\Theta}^{ref}_{k}\|^2_{F}\},
\end{aligned}
\end{equation*}
and
\begin{equation*}
\begin{aligned}
\mathop{\min}_{\tilde{\Theta}^{test}_{k}\in \mathbb{R}^{n\times (n+m)}} \{\|X^{test}_{k}-\tilde{\Theta}^{test}_{k}Z^{test}_{k}\|^{2}_{F}+\lambda \|\tilde{\Theta}^{test}_{k}\|^2_{F}\},
\end{aligned}
\end{equation*}
where $\lambda>0$ is a regularization parameter.
The solutions to the above problems are given by
\begin{equation} \label{ref}
\begin{aligned}
\hat{\Theta}_{k}^{ref}=\nu_{k}^{ref}(Z_{k}^{ref})^{T}(Z_{k}^{ref}(Z_{k}^{ref})^{T}+\lambda I_{n+m})^{-1},
\end{aligned}
\end{equation}
and 
\begin{equation}  \label{test}
\begin{aligned}
\hat{\Theta}_{k}^{test}=\nu_{k}^{test}(Z_{k}^{test})^{T}(Z_{k}^{test}(Z_{k}^{test})^{T}+\lambda I_{n+m})^{-1},
\end{aligned}
\end{equation}
 
respectively.
Consequently, an alarm is raised if the statistic $p_{k}\triangleq\|\hat{\Theta}_{k}^{ref}-\hat{\Theta}_{k}^{test}\|$ is greater than some threshold value. 

Notice that the above algorithm proposed in \cite{xin2023online} is based on linear systems. However, we conjecture that similar ideas will work for nonlinear systems, especially when the system operates around its equilibrium point. Alternatively, we can also leverage the learned deep Koopman linear representation in \cite{han2020deep} as the true representation of the system for the CPD algorithm. More specifically, one can replace the true state $\nu_{k}$ by the Koopman state $g(\nu_{k},\theta_\tau)$ (as will be defined in the next section).

On the other hand, although the paper \cite{xin2023online} presents a theoretical threshold that allows one to control the false alarm probability, the theoretical guarantees may not be accurate when nonlinearity exists. Hence, we use a different method to select the threshold value in this work. More specifically, we let $\bar{p}_{k}$ be the moving average of the time series $p_{k}$ using the most recent $W$ samples at each time step $k\geq 2N-1$. Then we raise an alarm if $p_{k}\geq c_{1}\bar{p}_{k}$, or equivalently, $\frac{p_{k}}{\bar{p}_{k}}\geq c_{1}$ for some constant $c_{1}>0$. Intuitively, the ratio  $\frac{p_{k}}{c_{k}}$ should be around $1$ when there is not a change point, and large when there is a change point. Consequently, a larger threshold $c_{1}$ corresponds to a lower false alarm rate and a lower true alarm rate. Further, if we approximate the time series $p_{k}$ as a sequence of i.i.d exponential random variables, we can use $c_{1}$ to represent a user-specified confidence parameter, e.g., $c_{1}=2$ corresponds to a confidence of around 0.865.

Note that due to the existence of nonlinearity, model uncertainty, and noise, one may want to use a large $c_{1}$ in the above detection rule to reduce the false alarm rate. Consequently, the above detection rule only aims to detect significant changes in model parameters. However, from a practical perspective, it is possible that one may want to detect small changes in model parameters as well. In particular, if a change that makes the current control law fail occurs, it should always be detected. In light of this, we also directly leverage tracking error as input data to the CPD algorithm. Under the condition that the old controller is able to achieve the goal/track the target in the nominal condition, a sequence of tracking errors that is amplifying over time would imply the presence of a change in system dynamics. More specifically, we again maintain two sliding windows of tracking error, each of length $N_{2}$, i.e., $E^{test}_{k}=\{e_{k}, e_{k-1}, \ldots, e_{k-N_{2}+1} \}$, $E^{ref}_{k}=\{e_{k-N_{2}}, e_{k-N_{2}-1},\ldots , e_{k-2N_{2}+1}\}$. Denote the average tracking error of $E^{test}_{k}$ as $\bar{E}^{test}_{k}$ and the average tracking error of $E^{ref}_{k}$ as $\bar{E}^{ref}_{k}$. We raise an alarm if $\bar{E}^{test}_{k}-E^{ref}_{k}>c_{2}$ for some positive constant $c_{2}$.

The overall algorithm is a composite of the parameter estimation-based CPD algorithm (the variant of \cite{xin2023online}), and the tracking error-based CPD algorithm presented above. We claim a change point if either one of the above CPD rules raises an alarm. 

\subsubsection{Cascade LQR controller Design}

We design a cascade control with a (i) Lyapunov-based controller using the known kinematics as the outer-loop that outputs the reference speed and yaw rate (denoted as $u_{ref}$ and $r_{ref}$, respectively) given the target pose and (ii) a deep Koopman-based linear quadratic regulator (LQR) with an integrator as an inner-loop that outputs the motor commands $a$ that tracks the reference speed and yaw rate.

The outer-loop is designed using Lyapunov theory and is based on the knowledge of the kinematics of the system \eqref{eq:asv_kin}. Note that the system is under-actuated and non-holonomic. In other words, the configuration of the thrusters do not allow us to directly control the lateral movement (along $v$ direction). To solve this problem and achieve station-keeping of the vehicle, we designed a switched control system such that the control strategy is changed when the vehicle is near the target.  Let $\eta_g = [x_g, y_g, \psi_g]^T$ be the target pose, and $d_g = \sqrt{(x - x_g)^2 + (y - y_g)^2}$ be the distance to the target. When the vehicle is far away from the target, i.e. $d_g \geq \bar d_g$, where $\bar d_g$ is the threshold distance, we design the outer-loop to control the distance error $d_g$ and the error between the yaw angle and the direction of the target position. On the other hand, when the vehicle is close to the target, i.e. $d_g < \bar d_g$, we design the outer-loop to control the distance to a virtual line that passes through the target position and is perpendicular to the yaw angle of the vehicle and the error between the yaw angle and the target yaw angle.

Let $\psi_{c} = \arctan((y - y_g)/(x - x_g))$ be the direction of the target position from the vehicle, and let $\eta_c = [x_g, y_g, \psi_c]$. Let the positive definite function $V_1(\eta) = \frac{1}{2}\|\eta - \eta_c \|^2$ be the Lyapunov candidate function for when $d_g \geq \bar d_g$.
From \eqref{eq:asv_kin}, the time derivative of $V_1$ is given as:
\begin{equation}
\begin{aligned}
    \frac{dV_1}{dt} &= (\eta - \eta_c)^T R(\psi)\nu \\
    &= ((x - x_g)\cos(\psi) - (y - y_g)\sin(\psi)) u \\
    &+ ((x - x_g)\sin(\psi) + (y - y_g)\cos(\psi)) v + (\psi - \psi_c)r
\end{aligned}
\end{equation}
We design the yaw rate $r_d$  and desired speed $u_d$ as follows:
\begin{align} \label{eq:outer1}
    r_d &= -k_\psi (\psi - \psi_c) \\
    u_d &= \frac{-k_x d_g^2 + ((x - x_g)\sin(\psi) + (y - y_g)\cos(\psi)) v}{(x - x_g)\cos(\psi) - (y - y_g)\sin(\psi)}
\end{align}
Assuming that the inner-loop follows the desired speed and yaw rate accurately, we let $u \approx u_d$ and $r \approx r_d$. Then,
\begin{equation} 
    \frac{dV_1}{dt} = -k_x (x - x_g)^2 - k_x (y - y_g)^2 -k_\psi (\psi - \psi_c)^2
\end{equation}
which is negative definite, and by the Lyapunov stability criterion, $V_1(\eta)$ is exponentially stable, or the position ($x$, $y$) of the vehicle converges to the target position ($x_g$, $y_g$) under this controller.
\begin{figure}[!t]
    \centering
    \includegraphics[width=0.8\linewidth]{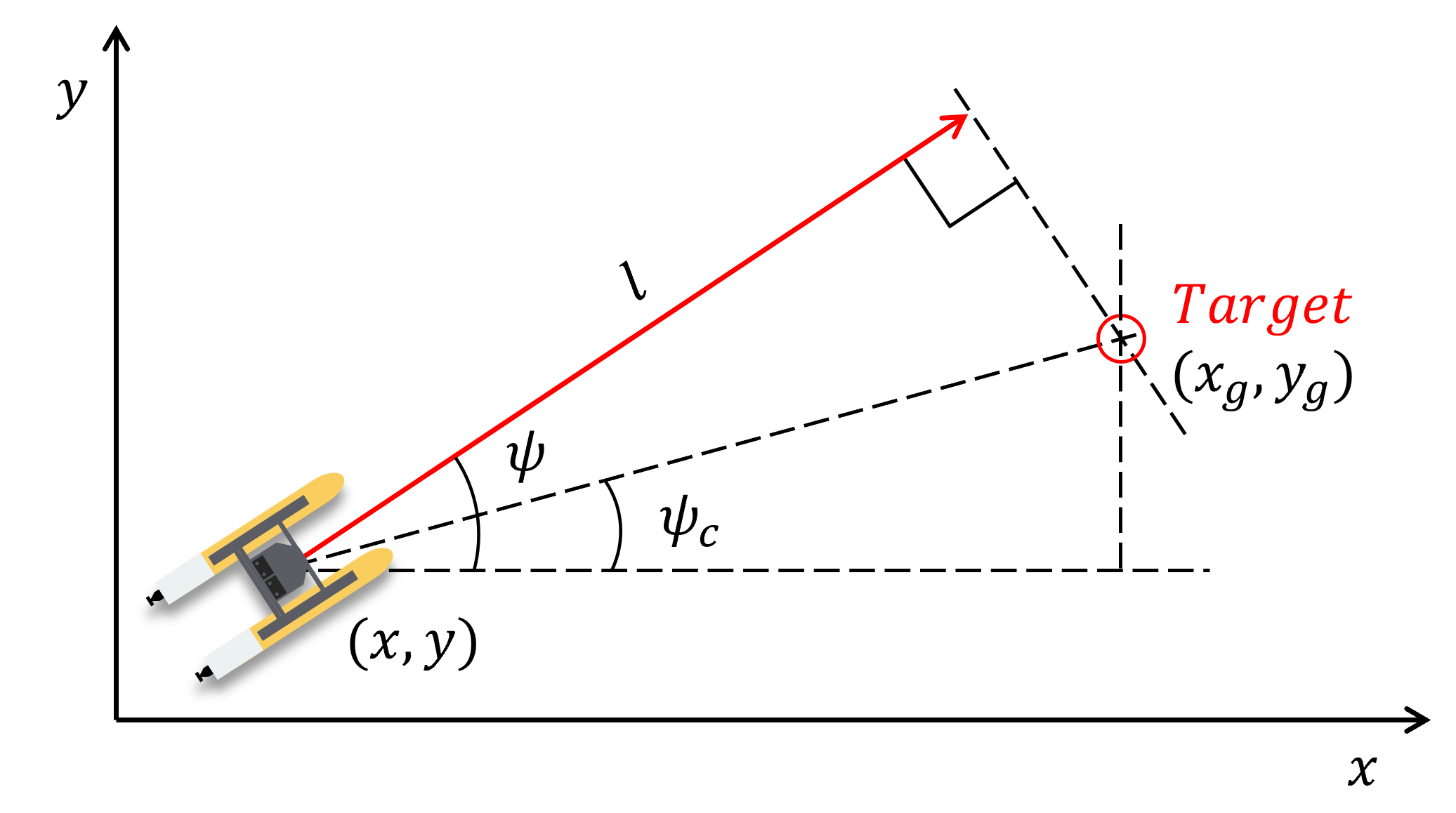}
    \caption{Diagram of the vehicle, the target, and distance to the virtual line perpendicular to the yaw angle}
    \label{fig:outer2_diagram}
\end{figure}
Once the vehicle is close to the target position, i.e., $d_g < \bar d_g$, we control the distance to a virtual line that passes through the target position and is perpendicular to the yaw angle of the vehicle. We also control the error between the yaw angle and the target yaw angle. The diagram of the virtual line and the distance $l$ is visualized in Fig. \ref{fig:outer2_diagram}, and $l$ can be written as
\begin{equation}
    l = (x - x_g)\cos(\psi - \psi_c) + (y - y_g)\sin(\psi - \psi_c)
\end{equation}
This distance can also be thought of as the ``along-track error" to the target position.
Let $V_2(\eta) = \frac{1}{2}(l^2 + (\psi - \psi_g)^2)$ be the Lyapunov candidate function for when $d_g < \bar d_g$. Then, from \eqref{eq:asv_kin}, the time derivative of $V_2$ is given as:
\begin{equation}
\begin{aligned}
    \frac{dV_2}{dt} &= l(u \cos(\psi) + v \sin(\psi)) \cos(\psi - \psi_c) \\
    &- l(x - x_g)\sin(\psi - \psi_c) (r - \dot\psi_c) \\
    &+l(u \sin(\psi) - v \cos(\psi)) \sin(\psi - \psi_c)\\ 
    &+ l(y - y_g)\cos(\psi - \psi_c) (r - \dot\psi_c) + r(\psi - \psi_g)
\end{aligned}
\end{equation}
where $\dot\psi_c = u\rho_u + v\rho_v$ and
\begin{equation*}
\begin{aligned}
    \rho_u &= \{(x - x_g)\cos(\psi) - (y-y_g)\sin(\psi)\}/d_g^2\\
    \rho_v &= \{(x - x_g)\sin(\psi) + (y-y_g)\cos(\psi)\}/d_g^2\\
\end{aligned}
\end{equation*}
Then, 
\begin{equation}
\begin{aligned}
    \frac{dV_2}{dt} &= lu\{\cos(\psi_c) + (x - x_g)\sin(\psi - \psi_c)\rho_u\\ 
    &- (y - y_g)\cos(\psi - \psi_c)\rho_v\} \\
    &+ lv\{\sin(\psi_c) + (x - x_g)\sin(\psi - \psi_c)\rho_v\\ 
    &- (y - y_g)\cos(\psi - \psi_c)\rho_u\} \\
    &+ lr\{(x - x_g)\sin(\psi - \psi_c) + (y - y_g)\cos(\psi - \psi_c)\} \\
    &+ r(\psi - \psi_g)
\end{aligned}
\end{equation}
We design the yaw rate $r_d$  and desired speed $u_d$ as follows: 
\begin{align} \label{eq:outer2}
    r_d &= -k_\psi (\psi - \psi_g) \\
    u_d &= -k_x l - \frac{N}{D}
\end{align}
where 
\begin{align*}
    N =& v\{\sin(\psi_c) + (x - x_g)\sin(\psi - \psi_c)\rho_v \\
    &- (y - y_g)\cos(\psi - \psi_c)\rho_u\} \\
    &+ r\{(x - x_g)\sin(\psi - \psi_c) + (y - y_g)\cos(\psi - \psi_c)\} \\
    D =& \cos(\psi_c) + (x - x_g)\sin(\psi - \psi_c)\rho_u\\
    &- (y - y_g)\cos(\psi - \psi_c)\rho_v
\end{align*}
Assuming that the inner-loop follows the desired speed and yaw rate accurately, we let $u \approx u_d$ and $r \approx r_d$. Then,
\begin{equation} 
    \frac{dV_2}{dt} = -k_x l^2 -k_\psi (\psi - \psi_g)^2
\end{equation}
which is negative definite, and by the Lyapunov stability criterion, $V_2(\eta)$ is exponentially stable. Therefore, using the controller in \eqref{eq:outer1} for $d_g \geq \bar{d}_g$, and the controller in \eqref{eq:outer2} for $d_g < \bar d_g$, the pose of the vehicle $\eta$ converges to the target pose $\eta_g$.

In our implementation, to account for the physical limit of the speed and the yaw rate,
we saturate the desired speed and yaw rate to obtain the reference speed and yaw rate:
\begin{equation}
u_{ref} = 
\begin{cases}
    u_{max}, & \text{if } u_d \geq u_{max}\\
    u_{min}, & \text{if } u_d \leq u_{min} \\
    u_d, & \text{otherwise}
\end{cases} 
\end{equation}
\begin{equation}
r_{ref} = 
\begin{cases}
    r_{max}, & \text{if } r_d \geq r_{max}\\
    r_{min}, & \text{if } r_d \leq r_{min}\\
    r_d, & \text{otherwise}
\end{cases} 
\end{equation}

We use the learned deep Koopman representation \eqref{eqq1} and \eqref{eqq2} to design the inner-loop, which is based on LQR with an integrator for tracking.

We augment the system in \eqref{eqq1} and \eqref{eqq2} with the following integrator,
\begin{equation}
    i_{k+1} = i_k + C_i\hat\nu_k - \nu^{ref}_k, \, i_0 = 0
\end{equation}
where $i_k \in \mathbb{R}^2$ is the integrator state at time step $k$, $C_i \in \mathbb{R}^{2\times3}$ is a projection matrix that selects the estimate of the vehicle speed and yaw rate, and $\nu^{ref}_k = [u^{ref}_k, r^{ref}_k]^T$ is the reference signal generated from the outer-loop. The augmented system is then
\begin{equation} \label{eq:aug_sys}
\tilde z_{k+1} = \tilde A \tilde z_k + \tilde B a_k - \tilde B^{ref} \nu^{ref}_k
\end{equation} where
$z_k = [i_k^T,\,z_k^T]^T$, 
\begin{equation}
    \tilde A = \begin{bmatrix}
        I & C_iC \\ 0 & A
    \end{bmatrix} \hspace{30pt}
    \tilde B = \begin{bmatrix}
        0 \\ B 
    \end{bmatrix} \hspace{30pt}
    \tilde B^{ref} = \begin{bmatrix}
        I \\ 0 
    \end{bmatrix}
\end{equation}
The cost function is chosen as
\begin{equation}\begin{aligned}
    J &= \sum_{k=0}^{\infty} i_k^T Q_i i_k + \hat\nu_k^T Q \hat\nu_k + a_k^TRa_k \\
    &= \sum_{k=0}^{\infty} \tilde z_k ^T \tilde Q z_k + a_k^TRa_k
\end{aligned}
\end{equation}
where 
\begin{equation}
    \tilde Q = \begin{bmatrix}
        Q_i & 0 \\ 0 & C^TQC
    \end{bmatrix} 
\end{equation}
We solve the discrete-time algebraic Riccati equation 
\begin{equation}
    P = \tilde A^T P \tilde A - (\tilde A^T P\tilde B)(R + \tilde B^T P \tilde B)^{-1} (\tilde B^T P \tilde A) + \tilde Q
\end{equation}
to obtain the feedback gain
\begin{equation}
    K = (R + \tilde B^T P \tilde B)^{-1} (\tilde B^T P \tilde A)
\end{equation}
and define the inner-loop control as
\begin{equation} \label{eq:outerloop}
    a_k = -K \tilde z_k
\end{equation}
Assuming that the solution to the Riccati equation exists for some positive-definite matrices $\tilde{Q}$ and $R$, the system in \eqref{eq:aug_sys} with the controller \eqref{eq:outerloop} is bounded-input bounded-output stable with respect to the reference input $\nu^{ref}_k$, and for a constant reference $\nu^{ref}_k$, $\hat{u}$ and $\hat{r}$ converge exponentially to $u^{ref}$ and $r^{ref}$.

\begin{figure}[!h]
\centering
    \includegraphics[width=0.5\textwidth]{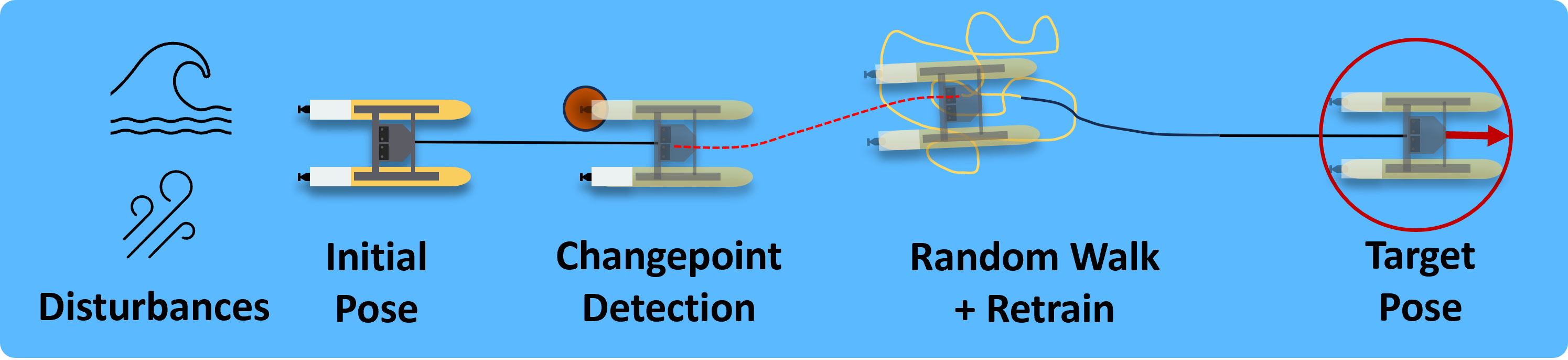}
\caption{\textcolor{black}{Overview of a station keeping test using the proposed framework.}}
\label{fig:wamv_overview}
\end{figure}

\section{RESULTS}\label{sec:results}

To test the system, station keeping tests have been designed. The test overview is shown in \ref{fig:wamv_overview}, and the black trajectory shows the movement of the ASV. The ASV begins from an initial pose under a nominal setting. During its approach to the target pose, the ASV may be impacted by environmental and/or system perturbations such as strong cross winds, waves, a change in mass, or compromised propulsion. This will cause a deviation from some reference path (red dotted trajectory). The CPD detects this change and triggers a data collection sequence for retraining the deep Koopman model. Once retrained, the ASV is able to converge to the target pose under optimal control (red solid arrow). 


\begin{figure}[htb]
    \centering
    \begin{subfigure}{0.15\textwidth}
        \includegraphics[width=\linewidth]{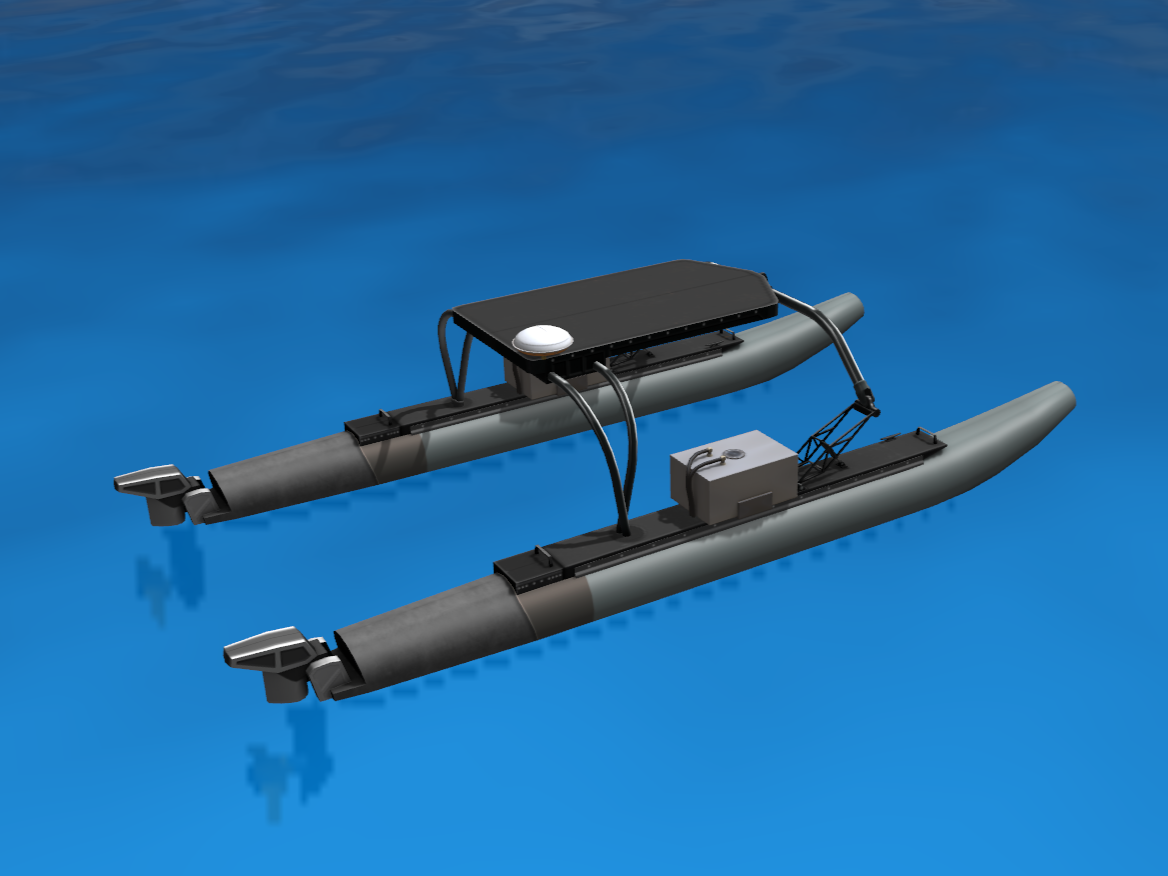}
        \caption{}
        \label{fig:wamv_sim_nom}
    \end{subfigure}\hfil
    \begin{subfigure}{0.15\textwidth}
        \includegraphics[width=\linewidth]{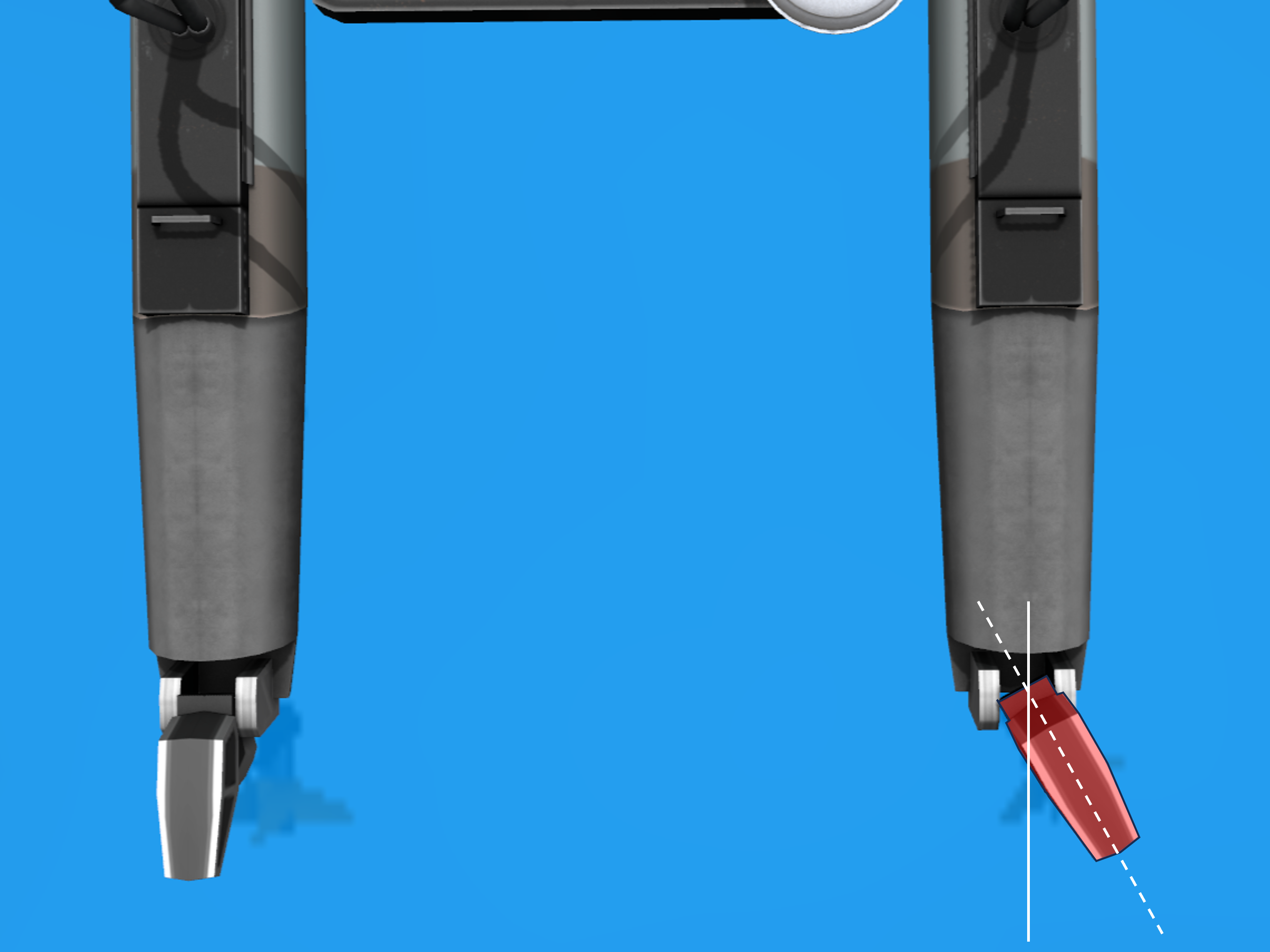}
        \caption{}
        \label{fig:wamv_rudder}
    \end{subfigure}\hfil
    \begin{subfigure}{0.15\textwidth}
        \includegraphics[width=\linewidth]{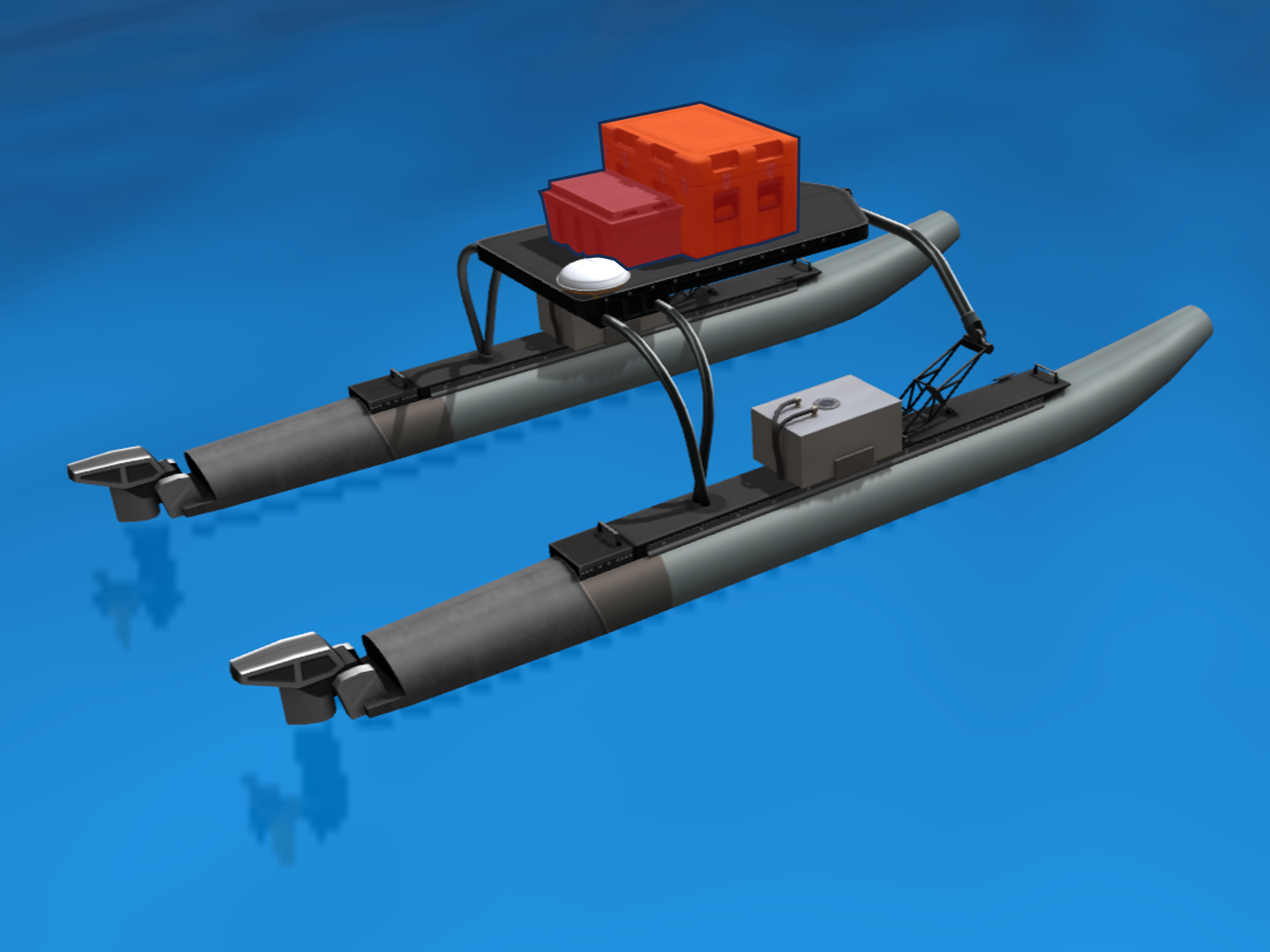}
        \caption{}
        \label{fig:wamv_sandbag}
    \end{subfigure}

    \medskip

    \begin{subfigure}{0.15\textwidth}
        \includegraphics[width=\linewidth]{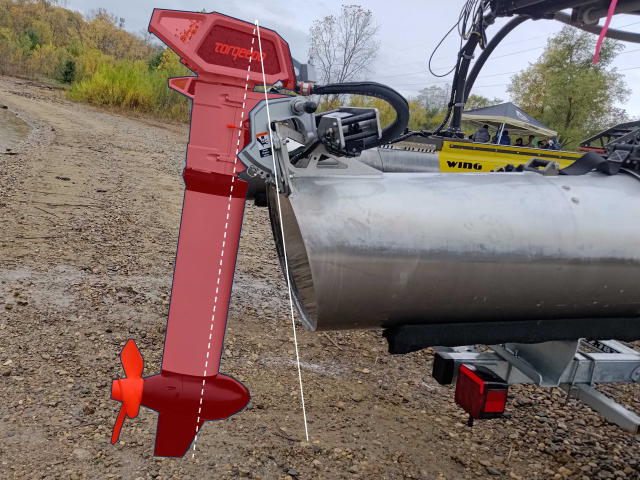}
        \caption{}
        \label{fig:wamv_trim}
    \end{subfigure}\hfil
    \begin{subfigure}{0.15\textwidth}
        \includegraphics[width=\linewidth]{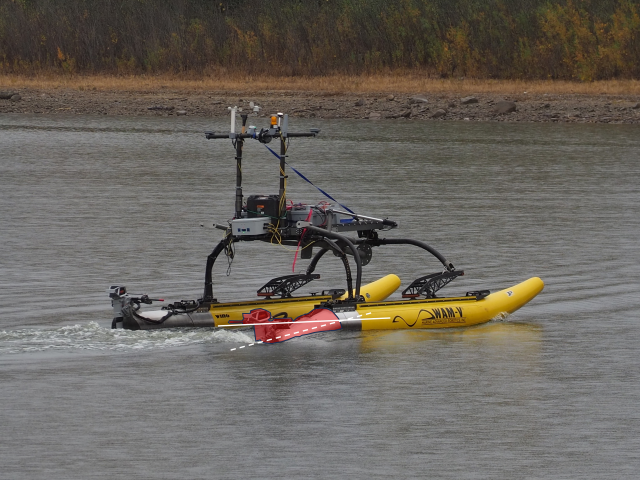}
        \caption{}
        \label{fig:wamv_trim_eff}
    \end{subfigure}\hfil
    \begin{subfigure}{0.15\textwidth}
        \includegraphics[width=\linewidth]{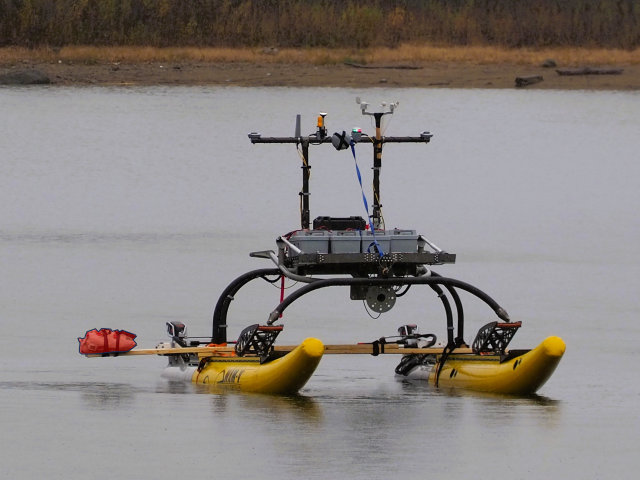}
        \caption{}
        \label{fig:wamv_sandbag}
    \end{subfigure}

    \caption{Nominal WAM-V 16 (a) and different changes that will affect system dynamics in simulation and the real world. (b) WAM-V 16 with a changed starboard rudder, (c) WAM-V 16 with a load on the platform (two boxes that weigh 100 kg), (d) and (e) WAM-V 16 with a changed starboard trim angle of $30^{\circ}$ and its effect. (f) WAM-V 16 with an unbalanced load (a 23 kg sandband) on the starboard side. The system changes have been highlighted in red.}
    \label{fig:wamv_changes}
\end{figure}

\subsection{Validation in Simulation}\label{sec:val_sim}

\begin{figure}[!h]
\centering
    \includegraphics[width=0.5\textwidth]{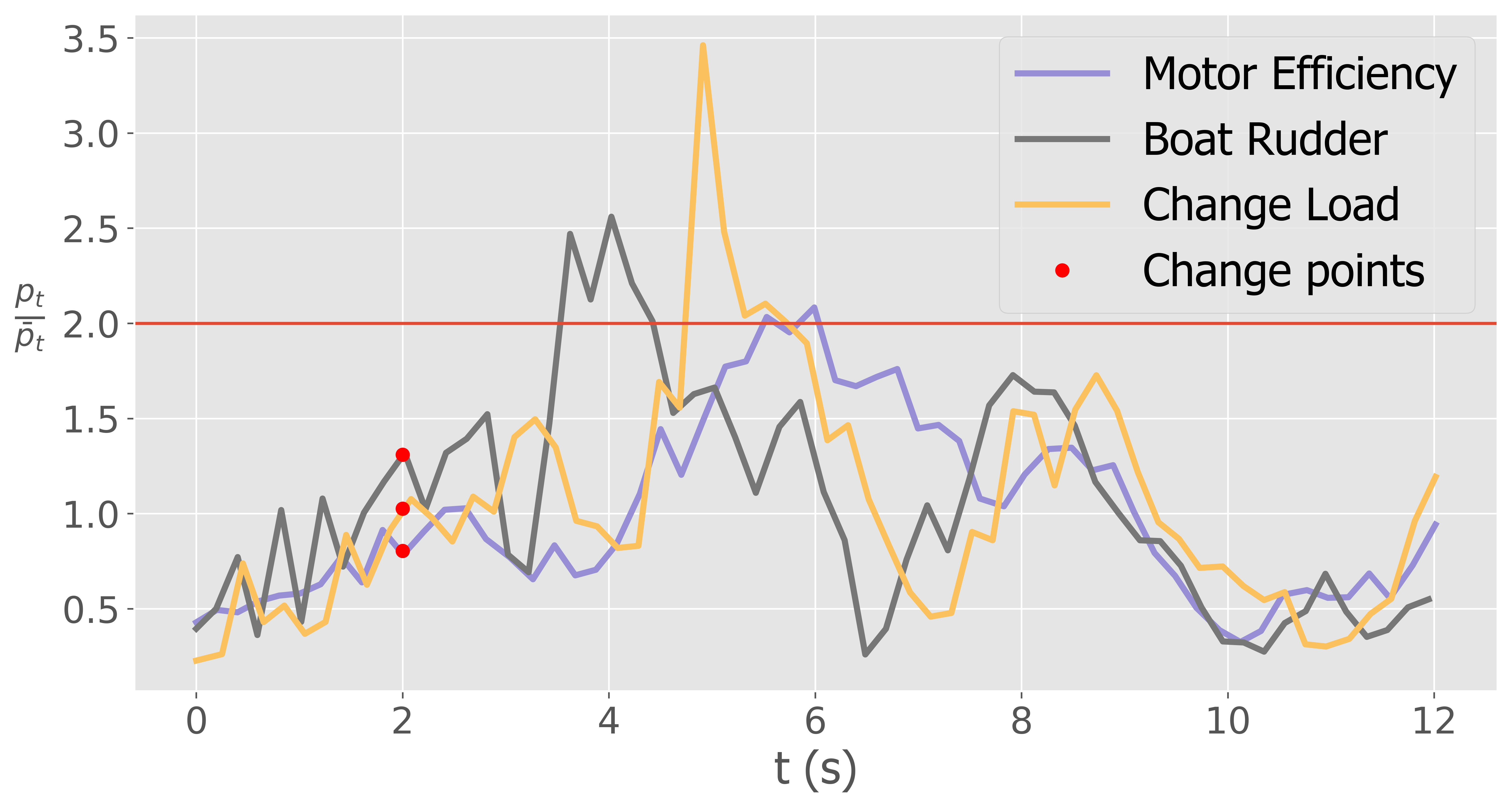}
\caption{Change point detector statistics. The red horizontal line represents the threshold $c = 2$. All change points happen at $t = 2s$.
}
\label{fig:cpd_sim}
\end{figure}

\begin{figure}[!h]
\centering
    \includegraphics[width=0.5\textwidth]{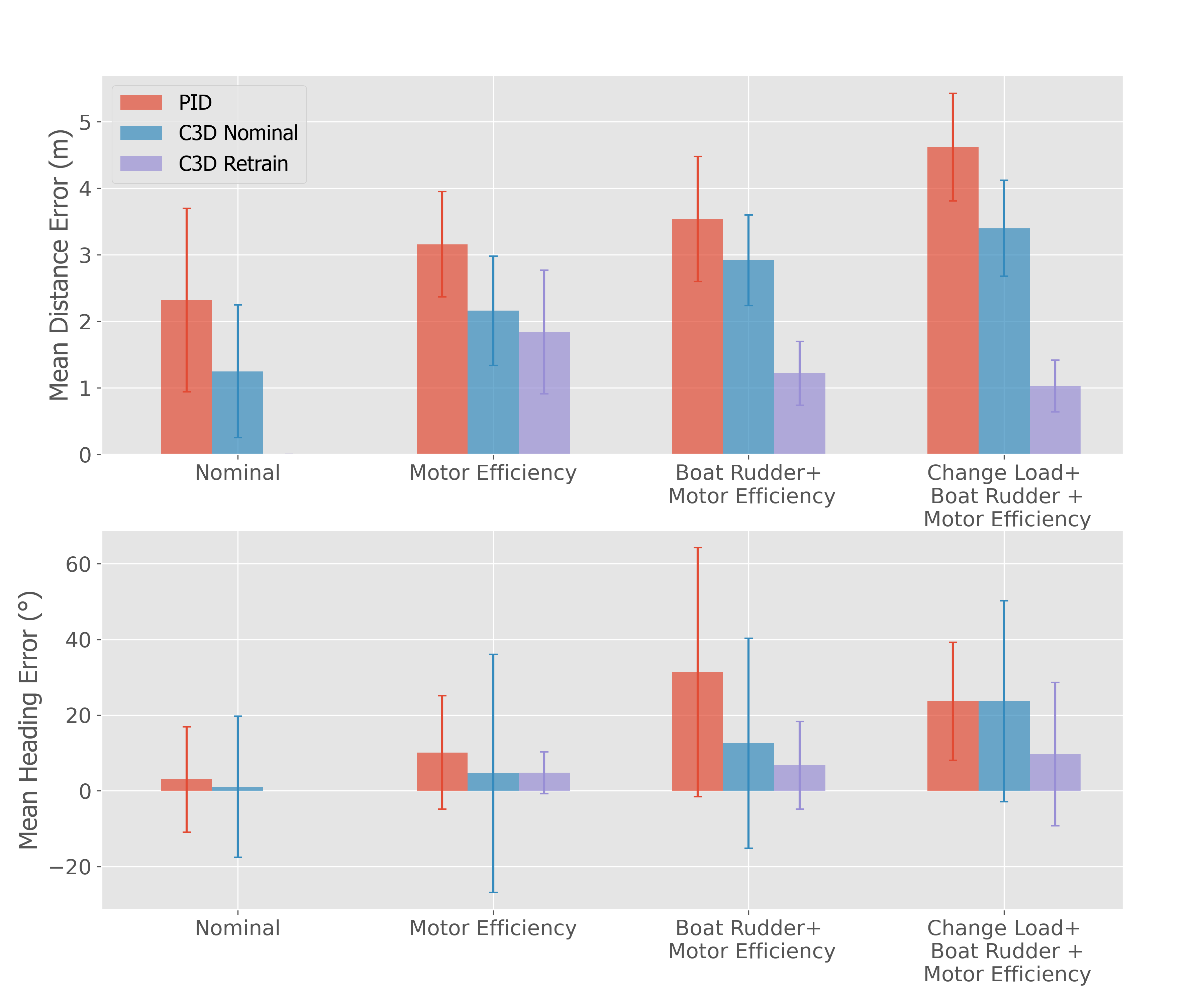}
\caption{Comparison of the controller for 4 tests in simulation. System changes have been labeled under the bars. The peaks of the bars represent the mean and the ends of the error bars depict $\pm 1$ standard error.
}
\label{fig:xtrack_bars_sim}
\end{figure}

The simulation that is used to test the overall station keeping strategy is implemented on Gazebo and the Robot Operating System (ROS) \cite{quigley2009ros}. In addition, plugins from the Virtual RobotX (VRX) simulator \cite{bingham19toward} are used to include the hydrodynamic-related components in the motion of the vehicle. In this architecture, Gazebo and Open Dynamics Engine (ODE) \cite{smith2005open} provide simulation of the rigid-body dynamics. To close the gap between the simulation and real world field experiments, motor delay, and sensor noises are also included in the simulation. These are important aspects observed during field experiments. We are also able to apply system changes, as discussed above, in simulation. Motor efficiency change is applied using a multiplier. Similarly, as shown in Fig \ref{fig:wamv_changes}, we can also apply rudder angle and load changes to impact the dynamics of the WAM-V 16. All of these changes can be triggered manually to test the change point and the controller.

To evaluate the control strategy for time-varying systems, 4 tests were carried out in the Gazebo simulator using different dynamics changes. The simulation validation is done on the desktop computer with a GeForce 2080 GPU and an Intel Core i7-8700 CPU. The first test has no system changes. The purpose of this test is to verify the performance of the baseline methods when the vehicle is operating under its nominal conditions, as shown in \ref{fig:wamv_sim_nom}. For the second test, the change is decreasing the right motor efficiency from 1 to 0.8. As for the third test, the changes are changing the angle of the right rudder by $18 \circ$ plus the decreased right motor efficiency. This will cause one motor to generate lateral force and make the vehicle deviate from the planned path. For the last test, a 100 kg load is added to the vehicle in addition to the right trim and right motor efficiency. This alters the mass and inertia of the vehicle and decreases its mobility.

To ensure the simulated vehicle is as close to the real one as possible, we apply sensor noise and motor delay that were measured from the real platform to the simulation. We assume the noises we inject into the simulation have a standard normal distribution. The standard deviation of linear position noise is $0.180$ m, the standard deviation of angular position noise is $0.008$ rad, the standard deviation of linear velocity noise is $0.013$ m/s, and the standard deviation of angular velocity noise is $0.008$ rad/s. 

Change point detection results are shown in \ref{fig:cpd_sim}. All three system changes have been detected while the nominal C3D is controlling the vehicle. It takes $3.6$ s to detect the right motor efficiency changing from $1$ to $0.8$, $1.8$ s to detect the angle of the right rudder from $0$ to $18^\circ$, and $2.8$ s to detect the added load. Once the change point is detected, the vehicle will be controlled by the random walk signal to collect new data and retrain the deep Koopman model. After the training, the system will automatically load the new model and keep performing the station keeping task.

\begin{figure}[!t]
\centering

\begin{subfigure}[b]{0.4\textwidth}
    \centering
        \includegraphics[width=\textwidth]{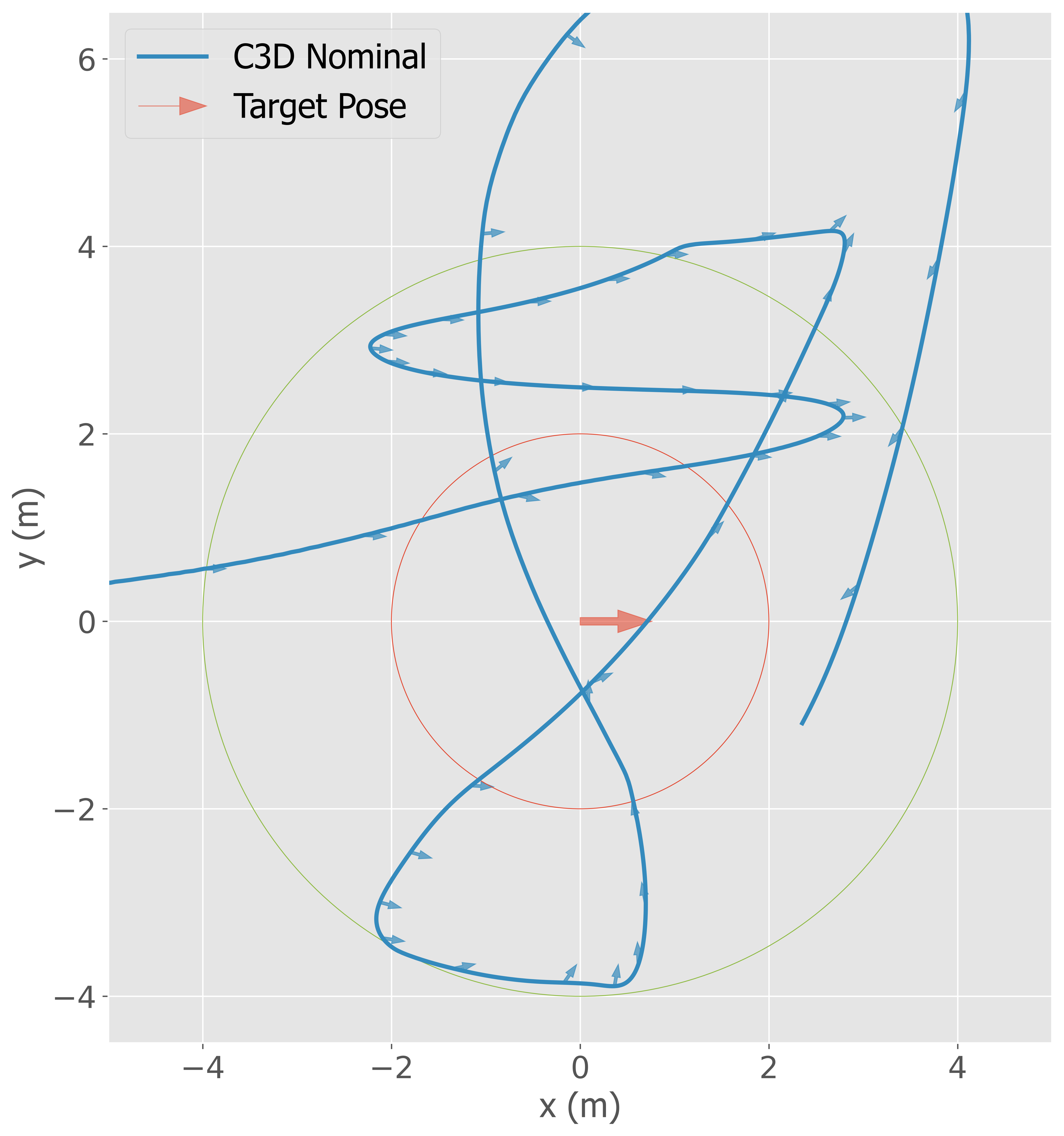}
    \caption{ }
    
\end{subfigure}
\begin{subfigure}[b]{0.4\textwidth}
    \centering
        \includegraphics[width=\textwidth]{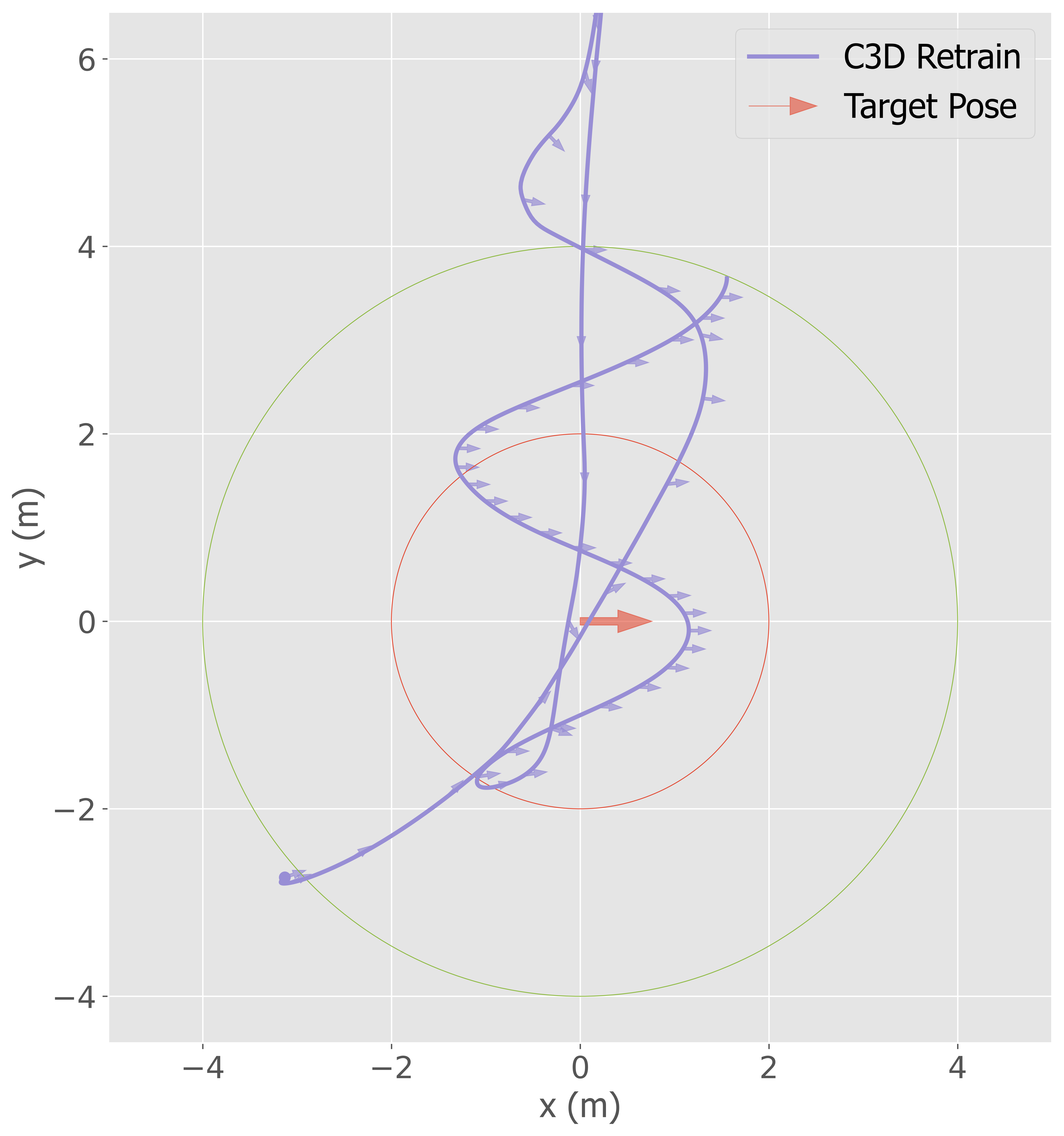}
    \caption{ }
\end{subfigure}
\caption{Trajectory plot of C3D before (a) and after (b) retrain for test 4 when right motor efficiency, right rudder, and increased weight have been applied. Each test lasts 200 seconds. The red circle and green circle represent areas that are within 2 and 4 m of the target position
respectively. }
\label{sim_traj}
\end{figure}
The performance of the station keeping strategy is evaluated not only using the magnitude of the distance and the heading errors ($e_d,e_h$) but also considering a steady behavior. For that reason, the standard deviation of the distance $s_d$ and heading $s_h$ are considered in the overall comparison. For a fair comparison, we begin recording the error once the vehicle is 2 meters away from the goal. Subsequently, we calculate the error statistics over the next 60 seconds. The results of the tests are summarized in Fig. \ref{fig:xtrack_bars_sim}. The graph shows the mean distance error (in meters) and mean heading error (in degrees) across various scenarios and control algorithms. Three algorithms, PID, C3D with the nominal model(C3D Nominal), and C3D with a retrained model (C3D Retrain), are evaluated under four distinct scenarios. Notably, the C3D Retrain algorithm consistently outperforms the others, minimizing both distance and heading errors, which is a position error of less than $1.9$ m and a heading error of less than $9.5^\circ$ in all cases. In summary, for these specific scenarios, C3D Retrain emerges as the most effective choice in terms of error reduction.

The trajectory plots can be found in Fig. \ref{sim_traj}. Unlike the trajectory of the C3D Nominal in blue, C3D Retrain trajectory exhibits a more direct and purposeful path toward the target (indicated by the red arrow). The trajectory avoids unnecessary loops and oscillations, suggesting a more efficient and optimized approach. The controller appears to be following a smoother trajectory, which could indicate improved control algorithms. Overall, the C3D Retrain's trajectory appears to be an improvement over the previous one and C3D with a retrained model achieves better performance than the rest of the controllers when controlling an underactuated ASV to reach a goal and keep the station.

\subsection{Experimental Validation}

A backseat-frontseat architecture was adopted to enhance the modularity and portability of our proposed controller. This is a common choice in marine vehicles. In our configuration, the change point detector, deep Koopman learning, and cascade controller reside in the backseat module, housing a Jetson Nano, while the real-time low-level controllers and sensors operate on the frontseat module, which is a Raspberry Pi 4. The backseat and frontseat components communicate through ROS, facilitating straightforward implementation. This architectural design allows flexibility of deployment across various ASV platforms such as BREAM (Boat for Robotic Engineering and Applied Machine Learning) \cite{lambert2020low}, irrespective of their underlying differences. 

A total of 60 hours were spent in Harner Lakes, IN for deployment, debugging, data collection, and fine-tuning. The final data was collected on October 25th between 11 am to 2 pm. The wind direction was southwest to south, and the wind speed was $2$ to $3$ m/s.

To evaluate the control framework, several tests were defined to observe its performance. The first test is to test the nominal system as shown in \ref{fig:wamv_real_nom} without any changes. For the second test, the change is decreasing the left motor efficiency from 1 to $0.8$. As for the third test, the changes are tilting the angle of the right engine by $30^\circ$ plus the decreased right motor efficiency shown in Fig. \ref{fig:wamv_trim}. This will cause one pontoon to be more submerged and generate more drag as shown in Fig. \ref{fig:wamv_trim_eff}. For the last test, on top of the right trim and right motor efficiency, a $23$ kg sandbag is added on the right side of the vehicle as shown in \ref{fig:wamv_sandbag}, which changes the mass and inertia of the vehicle and makes it more difficult to rotate.

\begin{figure}[!t]
\centering
    \includegraphics[width=0.5\textwidth]{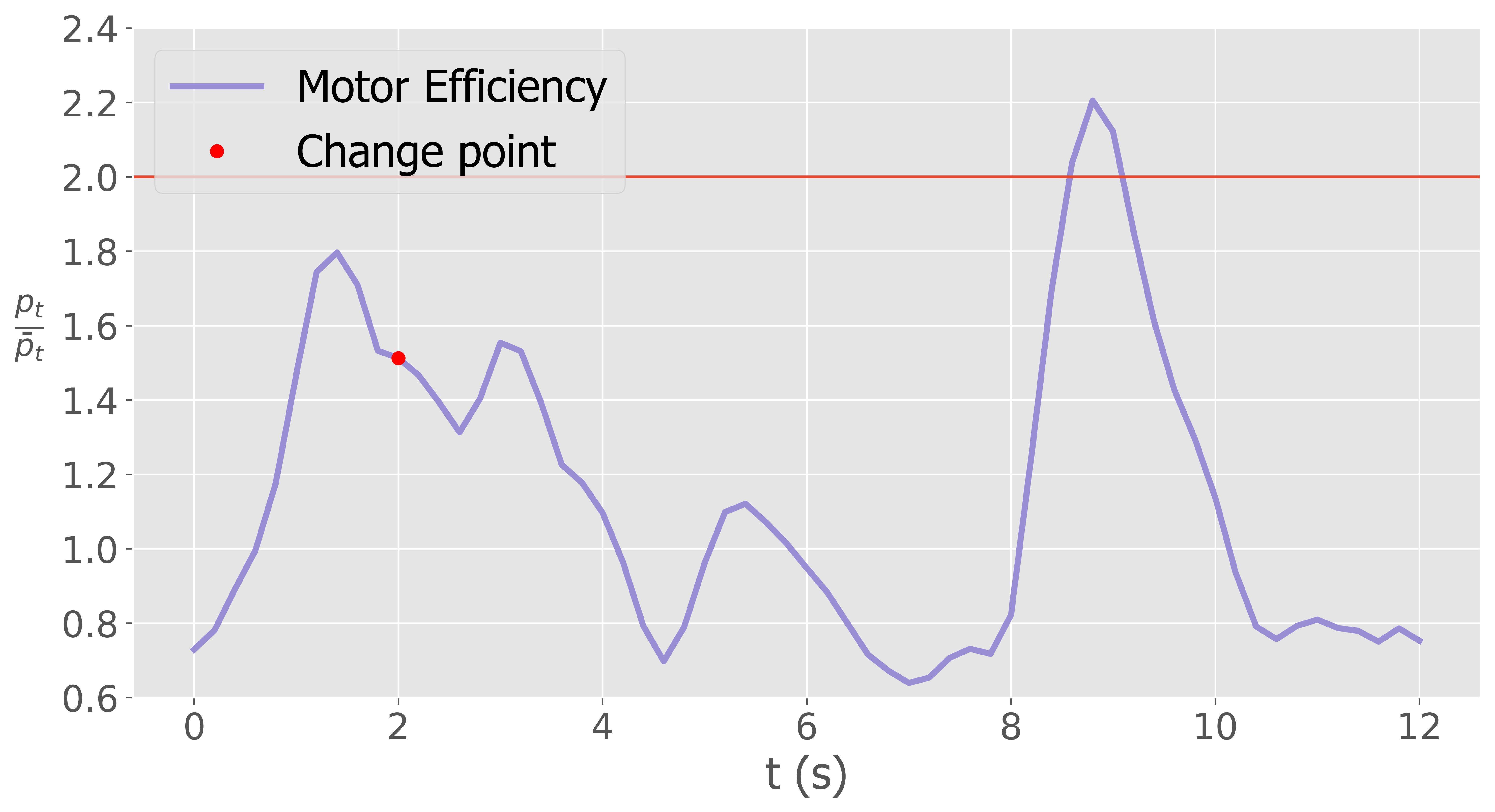}
\caption{Change point detector statistics during test 2. The red horizontal line represents the threshold $c = 2$. The motor efficiency is changed at $t = 2$ s. A change point is detected at $t=8.6$ s.
}
\label{fig:cpd_real}
\end{figure}

Motor efficiency change is applied during the test, while the right trim and the sandbag are applied when the WAM-V is on shore. The motor efficiency change is able to be detected by the change point detector online as shown in Fig. \ref{fig:cpd_real}, CPD successfully detects the change in motor efficiency from $1$ to $0.8$ after $6.6$ s. However, because the other changes (trim angle and unbalanced load) are implemented when the ASV is docked on the shore, retrain signals are manually sent to the deep Koopman learner after the ASV is launched. 


\begin{figure}[!t]
\centering
    \includegraphics[width=0.5\textwidth]{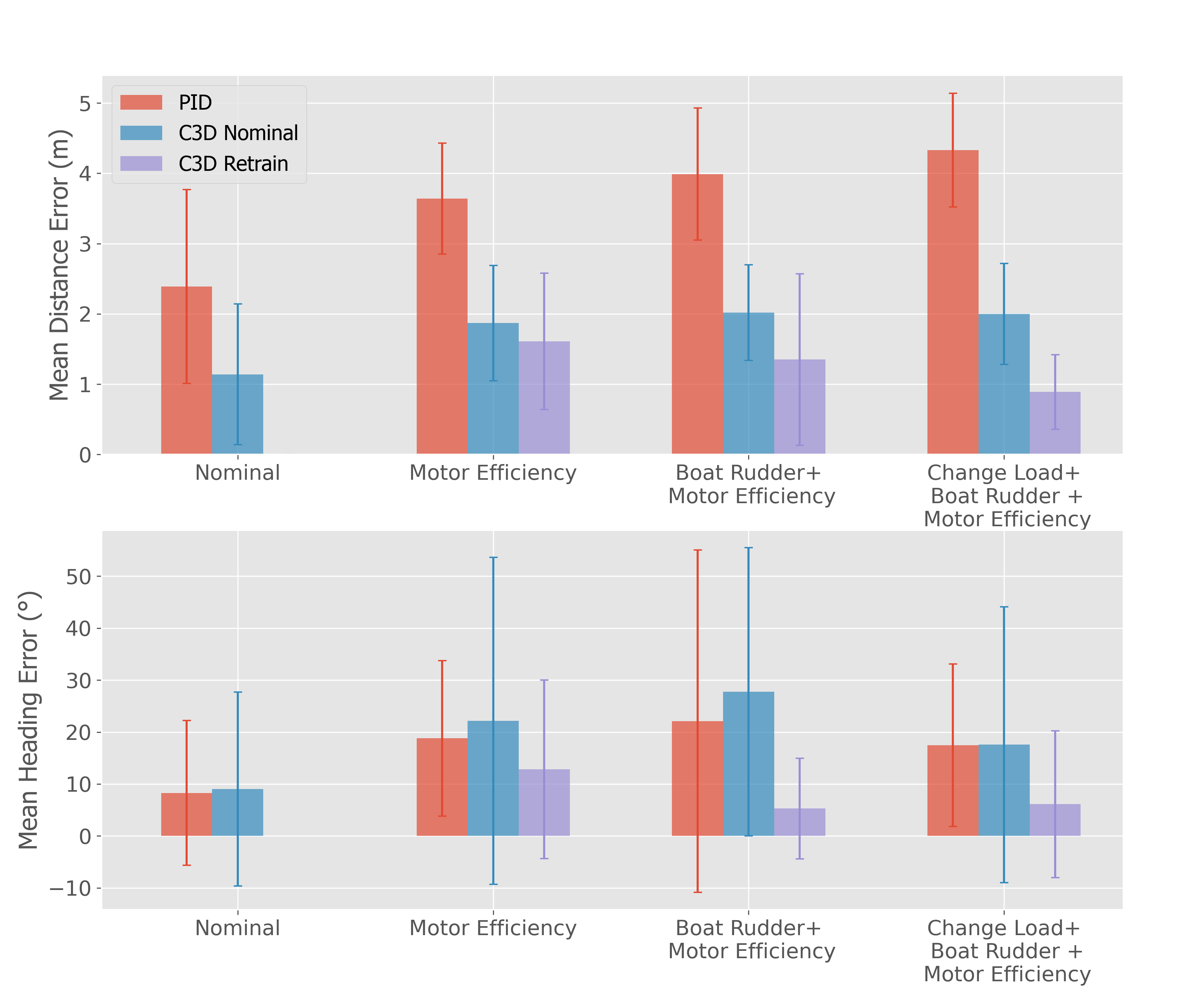}
\caption{Comparison of the controller for 4 tests in real-world experimental validation. System changes have been labeled under the bars. The peaks of the bars represent the mean, and the ends of the error bars depict $\pm 1$ standard error.
}
\label{fig:error_real}
\end{figure}

\begin{figure}[!h]
\centering

\begin{subfigure}[b]{0.5\textwidth}
    \centering
        \includegraphics[width=0.9\textwidth]{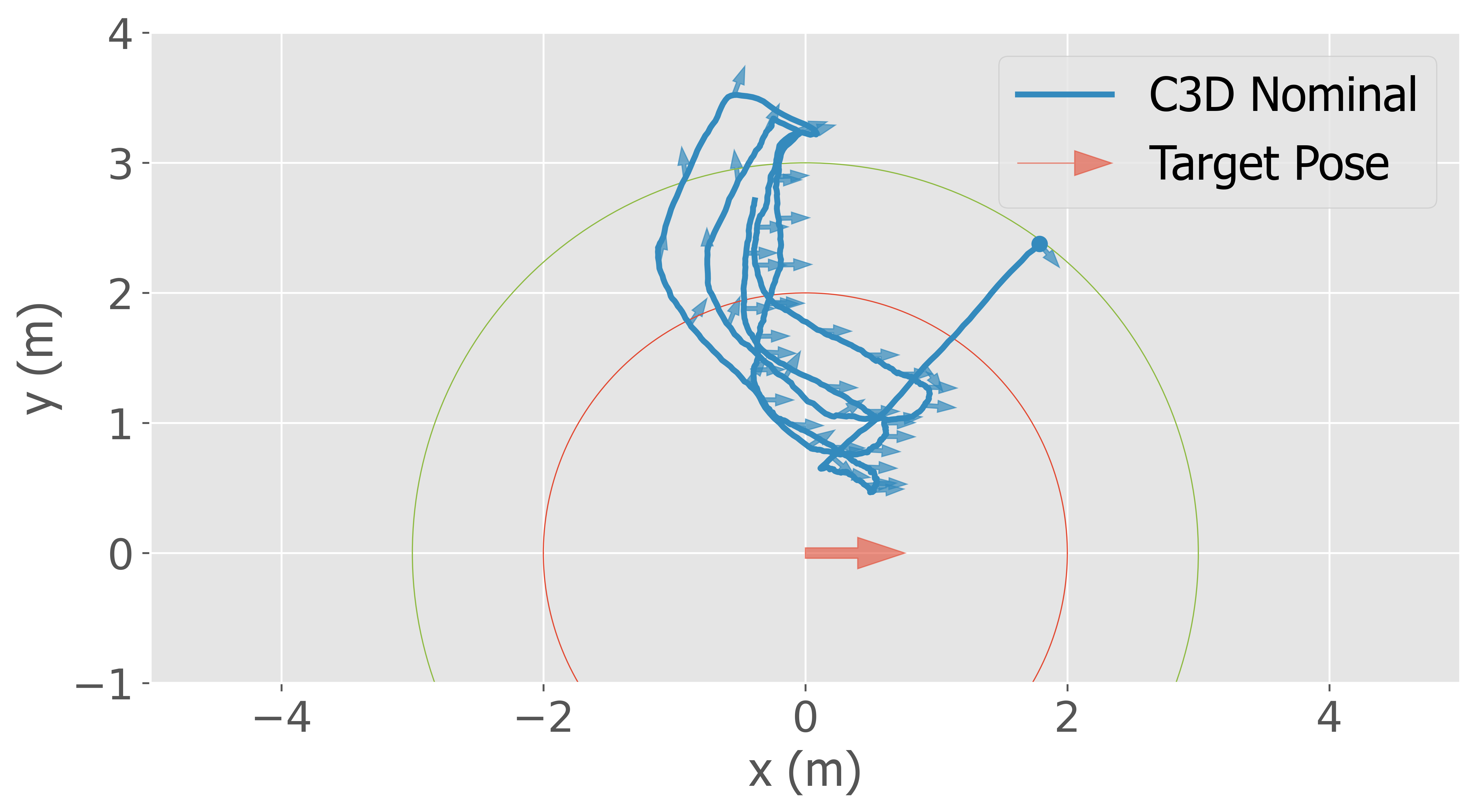}
    \caption{ }
    
\end{subfigure}
\begin{subfigure}[b]{0.5\textwidth}
    \centering
        \includegraphics[width=0.9\textwidth]{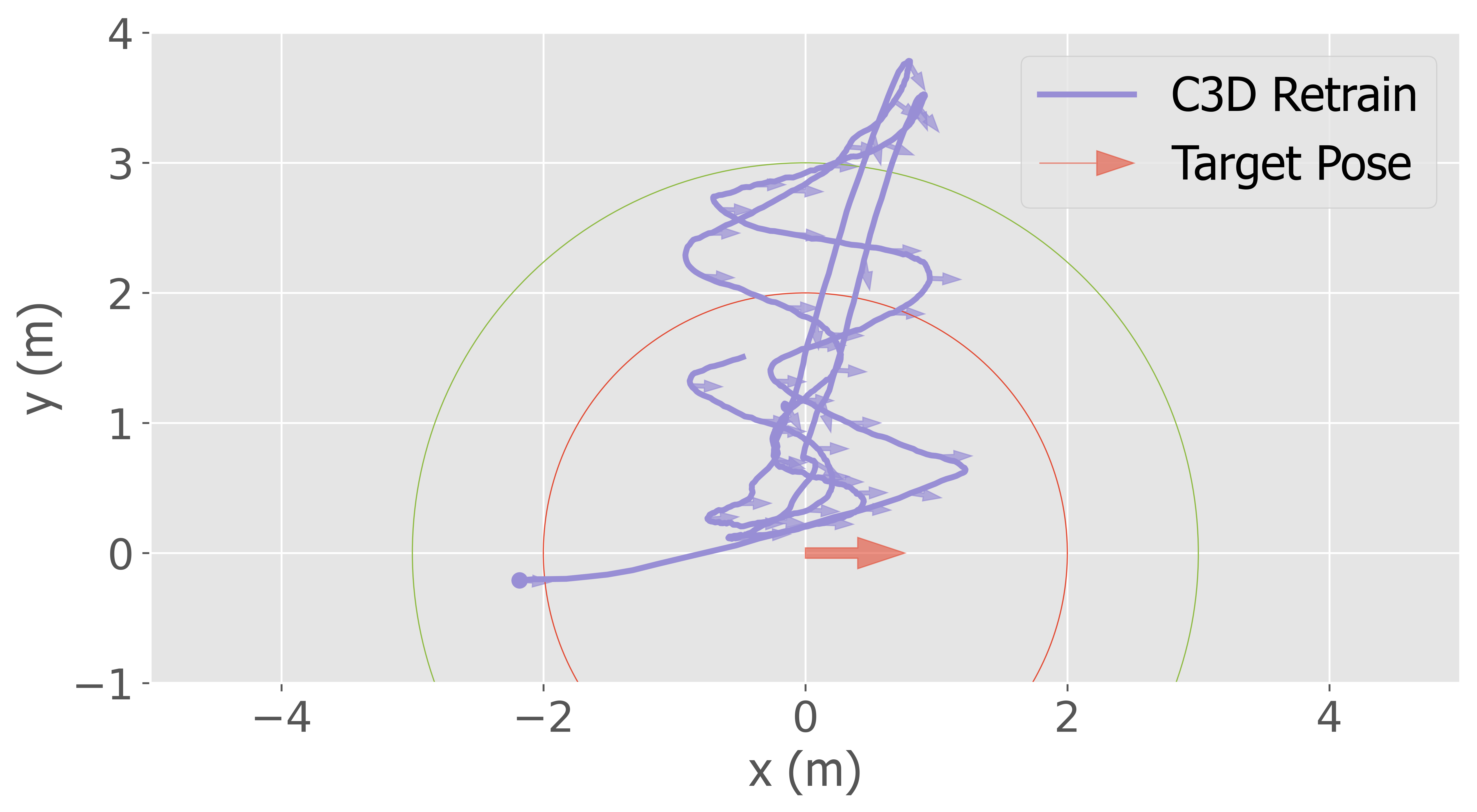}
    \caption{ }
    
\end{subfigure}
\caption{Trajectory plot of C3D before (a) and after (b) retrain for test 2 when right motor efficiency has been changed. Each test lasts 200 seconds. The red circle and green circle represent areas that are within 2 and 3 m of the target position respectively. }
\label{fig:traj_real}
\end{figure}

The mean distance error and heading error of the controller are shown in Fig. \ref{fig:error_real}. A total of 60 s are selected to calculate the mean and deviation. As different changes are applied to the system, the errors of the PID controller and the nominal C3D controller keep increasing. Due to the model change, the original PID parameters are no longer good enough and the pretrained nominal model is no longer accurate. After retraining, the proposed method's distance error even decreases as more changes have been applied, and it has a mean position error consistently below $1.6$ meters and a heading error below $12.8^\circ$ across all test cases.  As shown in Fig. \ref{fig:traj_real}, the C3D Nominal's trajectory exhibits a behavior where it returns to the target pose after drifting away. However, it initially points toward the goal position and then corrects the heading. In contrast, the C3D Retrain's trajectory achieves the same goal pose while maintaining a relatively small heading error. This zig-zag maneuver, similar to parallel parking, results in a smaller mean heading error.

In all cases, the C3D controller, after retraining, achieved minimal distance error and heading error in all tests. The nominal C3D controller also outperformed the PID controller. Though the environmental disturbances are slightly different from each test, C3D Retrain shows a more robust and effective performance. The results have also shown a successful sim-to-real transition. 

\section{CONCLSIONS}\label{sec:conc} 
In this paper, C3D, a framework incorporating a change point detector, deep Koopman learning, and a cascade controller is proposed and applied in ASV station keeping scenarios. To validate the performance of the proposed approach, baseline controllers such as PID and nominal cascade control are used for comparison through simulation and real-world experimental validations. 
 

The results obtained from the simulation clearly show the advantage of C3D with the retrained model over the other controllers, especially due to the accuracy of the updated neural network model in predicting the actual vehicle's state. In all the simulation test cases the proposed approach reaches a position error of less than $1.9$ m and a heading error of less than $9.7^\circ$. The C3D is deployed on an ASV platform, WAM-V 16, to evaluate its performance under real conditions. Data collection is carried out using GPS, compass, and IMU in field deployments to train a nominal model that accurately predicts the behavior of the ASVs. Three different model changes are applied to the WAM-V 16. In all the real-world tests, a mean position error consistently below $1.6$ meters and a heading error below $12.8^\circ$ across all test cases. 

Further development plans include improving the system model, implementing MPC, and considering reachability. A better system model can be obtained by including wind and wave disturbances in the state. Besides, it is beneficial to find a way to collect data more efficiently than the random walk. This would improve the ability of the C3D to better adapt to changes and generate optimal control commands. Additionally, it is worth exploring the use of MPC with C3D to improve control performance as MPC is robust to disturbances and uncertainties and can consider energy efficiency. Furthermore, reachability analysis can help the vehicle ensure that the system operates within safe states and avoids dangerous situations. Reachability analysis can also perform anomaly detection, which can trigger the retraining of the deep Koopman model and the adjustment of the control law.

In future work, this project will focus on enhancing the handling of vehicle damages and unforeseen conditions. Additionally, the framework will be extended to support multi-agent systems.


\section*{ACKNOWLEDGMENT}
This material is based upon work supported by the Defense Advanced Research Projects Agency (DARPA) via Contract No. N65236-23-C-8012 and under subcontract to Saab, Inc. as part of the RefleXAI project. Any opinions, findings and conclusions, or recommendations expressed in this material are those of the author(s) and do not necessarily reflect the views of the DARPA, the U.S. Government, or Saab, Inc.
\bibliographystyle{IEEEtran}
\bibliography{station_keeping.bib}
 
\end{document}